\documentclass[11pt]{article}
\usepackage{acl}
\usepackage{times}
\usepackage{latexsym}
\usepackage{booktabs}
\usepackage[T1]{fontenc}
\usepackage[utf8]{inputenc}
\usepackage{microtype}
\usepackage{inconsolata}
\usepackage{graphicx}
\usepackage{multirow}
\usepackage{amsmath}
\usepackage{enumerate}
\usepackage{enumitem}
\usepackage{subcaption}
\usepackage{longtable}
\usepackage[table]{xcolor}
\usepackage{listings}
\definecolor{myred}{RGB}{178, 34, 34} %
\definecolor{mygreen}{RGB}{34,139,34}   %
\definecolor{myred2}{RGB}{237, 211, 210} %
\definecolor{mygreen2}{RGB}{198, 232, 206} %
\definecolor{myblue2}{RGB}{218,232,252}
\definecolor{codegreen}{rgb}{0,0.6,0}
\definecolor{codegray}{rgb}{0.5,0.5,0.5}
\definecolor{codepink}{RGB}{252, 142, 172}
\definecolor{codepurple}{rgb}{0.58,0,0.82}
\definecolor{backcolour}{RGB}{245,245,245}
\definecolor{AuroraBlue}{HTML}{EAF3FB}
\definecolor{SkyBlue}{HTML}{BDE2F6}
\definecolor{SerenityBlue}{HTML}{73A5C6}
\definecolor{OxfordBlue}{HTML}{3373C4}
\definecolor{MidnightBlue}{HTML}{1E3F66}
\definecolor{LightGreen}{rgb}{0.88,1,0.88}
\definecolor{LightRed}{rgb}{1,0.88,0.88}

\definecolor{delim}{RGB}{20,105,176}
\definecolor{numb}{RGB}{106, 109, 32}
\definecolor{string}{rgb}{0.64,0.08,0.08}

\lstdefinestyle{jsonstyle}{
    showspaces=false,
    showtabs=false,
    breaklines=true,
    postbreak=\raisebox{0ex}[0ex][0ex]{\ensuremath{\color{gray}\hookrightarrow\space}},
    breakatwhitespace=true,
    basicstyle=\ttfamily\small,
    upquote=true,
    stringstyle=\color{string},
    literate=
     *{0}{{{\color{numb}0}}}{1}
      {1}{{{\color{numb}1}}}{1}
      {2}{{{\color{numb}2}}}{1}
      {3}{{{\color{numb}3}}}{1}
      {4}{{{\color{numb}4}}}{1}
      {5}{{{\color{numb}5}}}{1}
      {6}{{{\color{numb}6}}}{1}
      {7}{{{\color{numb}7}}}{1}
      {8}{{{\color{numb}8}}}{1}
      {9}{{{\color{numb}9}}}{1}
      {\{}{{{\color{delim}{\{}}}}{1}
      {\}}{{{\color{delim}{\}}}}}{1}
      {[}{{{\color{delim}{[}}}}{1}
      {]}{{{\color{delim}{]}}}}{1},
}

\lstdefinelanguage{json}{
    morestring=[b]", %
}

\lstdefinestyle{mystyle}{
    backgroundcolor=\color{backcolour},   
    commentstyle=\color{magenta},
    keywordstyle=\color{blue},
    numberstyle=\tiny\color{codegray},
    stringstyle=\color{codepurple},
    basicstyle=\fontfamily{\ttdefault}\footnotesize,
    breakatwhitespace=false,         
    breaklines=true,                 
    keepspaces=true,    
    frame=single,
    numbersep=5pt,                  
    showspaces=false,                
    showstringspaces=false,
    showtabs=false,                  
    tabsize=2,
    classoffset=1, %
    keywordstyle=\color{violet},
    classoffset=0,
}
\lstdefinelanguage{JavaScript}{
  keywords={typeof, new, true, false, catch, function, return, null, catch, switch, var, if, in, while, do, else, case, break},
  keywordstyle=\color{blue}\bfseries,
  ndkeywords={class, export, boolean, throw, implements, import, this},
  ndkeywordstyle=\color{darkgray}\bfseries,
  identifierstyle=\color{black},
  sensitive=false,
  comment=[l]{//},
  morecomment=[s]{/*}{*/},
  commentstyle=\color{purple}\ttfamily,
  stringstyle=\color{red}\ttfamily,
  morestring=[b]',
  morestring=[b]"
}

\title{InsightEval: An Expert-Curated Benchmark for Assessing Insight Discovery in LLM-Driven Data Agents}

\author{
\textbf{Zhenghao Zhu\textsuperscript{1,*}},
\textbf{Yuanfeng Song\textsuperscript{2,*}},
\textbf{Xing Chen\textsuperscript{2}},
\textbf{Chengzhong Liu\textsuperscript{1}},
\\
\textbf{Yakun Cui\textsuperscript{1}},
\textbf{Caleb Chen Cao\textsuperscript{1,$\dagger$}},
\textbf{Sirui Han\textsuperscript{1,$\dagger$}},
\textbf{Yike Guo\textsuperscript{1}}
\\
\textsuperscript{1}The Hong Kong University of Science and Technology, Hong Kong, China\\
\textsuperscript{2}ByteDance, China \\
}

\begin{document}
\maketitle

\renewcommand{\thefootnote}{\fnsymbol{footnote}}
\footnotetext[1]{These authors contributed equally to this work.} 
\footnotetext[2]{Corresponding authors.} 
\renewcommand{\thefootnote}{\arabic{footnote}}

\begin{abstract}

Data analysis has become an indispensable part of scientific research. To discover the latent knowledge and insights hidden within massive datasets, we need to perform deep exploratory analysis to realize their full value. With the advent of large language models (LLMs) and multi-agent systems, more and more researchers are making use of these technologies for insight discovery. However, there are few benchmarks for evaluating insight discovery capabilities. As one of the most comprehensive existing frameworks, InsightBench also suffers from many critical flaws: format inconsistencies, poorly conceived objectives, and redundant insights. These issues may significantly affect the quality of data and the evaluation of agents. To address these issues, we thoroughly investigate shortcomings in InsightBench and propose essential criteria for a high-quality insight benchmark. Regarding this, we develop a data-curation pipeline to construct a new dataset named InsightEval. We further introduce a novel metric to measure the exploratory performance of agents. Through extensive experiments on InsightEval, we highlight prevailing challenges in automated insight discovery and raise some key findings to guide future research in this promising direction. 

\end{abstract}

\section{Introduction}

In a data-driven world, it is increasingly significant to understand and interpret vast and structured datasets \cite{qin2025multitend,lu2025bridging}. To uncover meaningful insights, data analysts require not only considering the apparent information, but also summarizing deeper patterns and relationships embedded within the dataset. Before the era of large language models (LLMs), common approaches largely relied on data processing libraries such as Pandas, NumPy, and Jupyter Notebook~\cite{yin2023natural}. Therefore, insight discovery is primarily dependent on extensive manual analysis and specialized domain knowledge. Currently, LLMs have promoted the development of agent-based systems for automated data analysis and insight extraction. Recent works such as InsightPilot~\cite{ma2023insightpilot} and InsightLens~\cite{weng2025insightlens} enable interactive data exploration via natural language, helping users rapidly identify key information.

However, comprehensive benchmarks for evaluating agents' insight exploration capabilities remain insufficient. As one of the few publicly available datasets in this domain, InsightBench~\cite{sahuinsightbench} exhibits significant flaws and inconsistencies, underscoring the need for a higher‑quality, more comprehensive benchmark for insight discovery. Therefore, we performed a thorough analysis of the existing Insight dataset and found numerous latent issues and inconsistencies. For example, the dataset contains substantial missing information, and some goal definitions are poorly conceived and overly broad. Several questions mention features or column names that are not in the source tables. In addition, the evaluation framework to assess agent performance is not sufficiently comprehensive, as it overlooks the accuracy of the generated insights and the novelty of the capability.

To address the issues mentioned above, we performed a deep analysis of existing deficiencies and designed three essential criteria for a high‑quality insight benchmark. Moreover, we suggest two new insight types, Evaluative and Exploratory. Guided by these principles, we designed a dataset construction pipeline: (1) Refine the original goal. (2) Verify existing questions and generate new questions. (3) Answer questions and generate insights. (4) Summarize insights. Through the pipeline, we constructed InsightEval, a novel benchmark comprising 1000 insights covering six types, outperforming existing benchmarks like InsightBench. We enforced rigorous quality controls by combining both automated checks and expert review.

In the evaluation phase, the previous assessment only relies on a single biased evaluator while ignoring erroneous outputs, and cannot recognize novel insights. Therefore, we adopted insight recall and insight precision measurement, and proposed a new metric called Insight $F1$ to comprehensively assess the agent's ability to discover insights. Furthermore, we introduced a novel metric to evaluate the ability to uncover previously unannotated insights.

We benchmarked numerous baselines, including popular models and agent frameworks.
Our results reveal that the Insight $F1$ Score can better reflect the agent's insight discovery ability, and our InsightEval dataset provides a more comprehensive and deeper assessment.

In summary, our contributions are as follows:
\begin{itemize}
    \item We conduct a rigorous analysis of existing insight benchmarks and define the requirements for a high‑quality insight dataset.
    \item We introduce InsightEval, a new benchmark specifically designed to assess agents’ data analysis and insight discovery capabilities.
    \item We propose a comprehensive evaluation framework that combines Insight $F1$ metrics and novelty measurement.
    \item Our findings demonstrate that our InsightEval and evaluation framework can provide a comprehensive, in-depth, and accurate assessment of insight discovery capability.
\end{itemize}

\begin{table*}[t!]
\small
\centering
\fontsize{8.5pt}{8.5pt}\selectfont 
\renewcommand\arraystretch{1.2}
\setlength{\tabcolsep}{2mm}

\begin{tabular}{p{0.08\textwidth} p{0.60\textwidth} p{0.25\textwidth}}

\toprule
\textbf{Error Type}  & \textbf{Examples} & \textbf{Explanation} \\
\midrule

Ambiguous Goal \newline (\textbf{E1}) &   
\textbf{Goal}: Find the discrepancy and imbalance in the distribution of incidents assigned across categories. (\textit{Flag-1})
\newline
\textbf{Goal}: Identify trends and underlying factors or correlations contributing to the increase in TTR. (\textit{Flag-10}) &
These Goals are overly vague and non-actionable, lacking explicit metric definitions, analysis dimensions, and temporal scope. \\

\midrule

Undefined Data Type \newline (\textbf{E2}) & 
\textbf{Question}: Do we observe any trend in the volume of incidents? \newline
\textbf{Data Type}: time series (\textit{Flag-4})\newline
\textbf{Question}: How does the success rate of goals across different categories compare?  \newline
\textbf{Data Type}: comparative (\textit{Flag-35}) & 
The predefined four categories are Descriptive, Diagnostic, Predictive, and Prescriptive. The example shows undefined types. \\

\midrule

Erroneous Questions \newline (\textbf{E3}) &   
\textbf{Question}: How do the distributions ... compare across departments? (\textit{Flag-80})
\newline
\textbf{Table Columns Name}: category, state, closed at, opened at, closed by, number, sys updated by, location, assigned to, caller id, sys updated on, short description, priority, assignment group &  
The questions raised do not match the content of the table and involve non-existent column names. \\

\midrule

Irrational \newline Insights \newline (\textbf{E4}) &   
\textbf{Insight}: Insufficient data to identify trends in resource allocation for 'Cost Reduction' goals. (\textit{Flag-29})
\newline
\textbf{Table Schema}: Column: category (object), ... values: ['Employee Satisfaction', 'Cost Reduction', 'Efficiency', 'Customer Satisfaction', 'Revenue Growth'] & 
This insight states that the data is insufficient for analysis. However, there is a corresponding "Cost Reduction" in the table schema. \\

\midrule

Reduplicative Insights \newline (\textbf{E5}) &   
\textbf{Insight 1}: Specific hardware issues related to Printer Malfunctioning are predominantly mentioned in incident descriptions. (\textit{Flag-1})
\newline
\textbf{Insight 2}: Most of the hardware incidents are related to printer issues. (\textit{Flag-1})
&
These two insights both state that hardware issues mainly revolve around printer malfunctions. \\

\bottomrule
\end{tabular}
\caption{Issues and Deficiencies in InsightBench. \textit{Flag-1}, \textit{Flag-10} etc. means the dataset name in InsightBench.}
\label{InsightBenchIssue}
\end{table*}

\section{Related Work}

\paragraph{Data Analytical Agents.}
Several agent systems and frameworks have been proposed in the data analysis field. In data visualization, MatPlotAgent~\cite{yang2024matplotagent} combines code and multimodal LLMs, nvAgent~\cite{ouyang2025nvagent} and MultiVis-Agent~\cite{lu2026multivis} generate executable programming code to visualize data. For exploratory insight, InsightPilot~\cite{ma2023insightpilot} and InsightLens~\cite{weng2025insightlens} leverage analytic selection and multi‐agent dialogue extraction to help users uncover and organize insights via natural‐language interaction effectively. DAgent~\cite{xu2025dagent} and an LLM‑based SQL‐generation approach~\cite{perez2025llm} generate SQL over databases to extract information and synthesize textual reports and insights. InsightBench~\cite{sahuinsightbench} proposes Agent Poirot, a multi‐agent framework that iteratively generates questions, produces executable code, and derives insights. Other work includes the LangChain Pandas framework~\cite{langchainpandas}, AutoGen~\cite{wu2024autogen} for customizable agent orchestration, and the ReAct~\cite{yao2023react} prompting strategies, which use reasoning and actions to enhance LLM decision‐making capabilities. 

\paragraph{Data Science Benchmarks.}
In Text‑to‑SQL research, Spider 2.0~\cite{leispider} and EHRSQL~\cite{lee2022ehrsql} introduce multi‑step query workflows in general and clinical contexts, including temporal and unanswerable queries. However, NL2SQL‑BUGs~\cite{liu2025nl2sql}, VisEval~\cite{chen2024viseval}, and PRACTIQ~\cite{dong2025practiq} address semantic‑error detection, visualization, conversational and ambiguous query handling, respectively. In the code‑generation and tabular‑analysis domains, DS‑1000~\cite{lai2023ds} and JuPyT5~\cite{chandel2022training} supply real‑world programming tasks from Stack Overflow and Jupyter notebooks paired with test-execution and DSP‑based evaluation. In the area of tabular data analysis, InsightBench~\cite{sahuinsightbench} and InfiAgent‑DABench~\cite{hu2024infiagent} focus on insight generation, covering query formulation, answer parsing, and summarization. They also assess LLM‑based agent performance through structured prompting for automated evaluation.

Our dataset provides a comprehensive and rigorous benchmark for insight discovery.
It has the following key characteristics: accuracy, clarity, and comprehensiveness. Experiments have clearly shown the reliability and high quality of this dataset. We sincerely hope that this dataset can promote the development of the entire field.

\section{Error Analysis of Existing Insight Datasets}

\subsection{Insight Discovery Task Formulation}
In the insight discovery task, current methods adopt a multi-agent architecture. Given tabular data and a predefined goal as inputs, the agent then generates a series of questions aligned with the goal and resolves these questions to produce answers. Subsequently, insights are discovered from the answers and are finally synthesized into a summary. There are two advanced and distinct paradigms for obtaining key information from data when solving the questions. Pérez et al.~\cite{perez2025llm} uses SQL to extract information, while Agent Poirot \cite{sahuinsightbench} employs Python code and data analysis libraries to retrieve the vital information.

\begin{figure}
    \centering
    \includegraphics[width=0.9\linewidth]{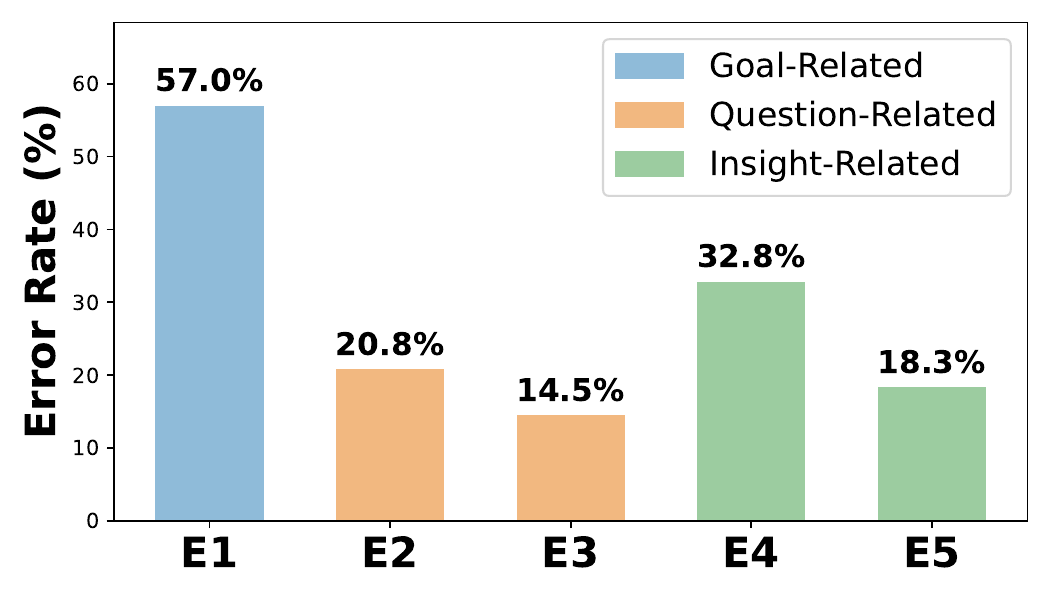}
    \caption{The proportion of each error type. E1, E2, E3, E4, E5 correspond to the Error Type in Table \ref{InsightBenchIssue}.}
    \label{fig:rawdata_issue}
\end{figure}

\subsection{Preliminary Study}

To establish a reliable and higher-quality benchmark, we conducted an in-depth preliminary study of the current InsightBench dataset~\cite{sahuinsightbench}. To assess quality, we utilized GPT-4o~\cite{openai2024gpt4ocard} as the backbone of the multi-agent framework called Agent Poirot, randomly sampling 50 data points and generating insights. Through comprehensive analysis, we uncover numerous issues and deficiencies that undermine both data quality and the evaluation process. We list these problems in Table~\ref{InsightBenchIssue} and also calculate the proportion of each type of problem in the dataset, which is shown in Figure~\ref{fig:rawdata_issue}. Finally, we summarize four key observations as follows:

\paragraph{Observation 1: Dataset Formatting and Textual Errors.}
We identified multiple examples of missing or flawed content in InsightBench. Specifically, some lacked a task goal, while others were missing their associated tabular data. We consider that these omissions resulted from oversights during the dataset construction. Moreover, we encountered cases in which a question and its answer were present, but the corresponding analytical insight was omitted. Meanwhile, several questions even contained insight types that were not defined. These inconsistencies may cause deviations in the subsequent analysis and evaluation results.

\paragraph{Observation 2: Ambiguous and Overly Broad Goals.} 
Among those samples, we observed that many goals were overly broad and lacked specificity. Note that a table can provide numerous potential insights through the comparative analysis of different columns. Consequently, a goal defined too generically often yields insights that are diffuse, unfocused, and lack substantive value.

\paragraph{Observation 3: Substandard Quality of Generated Questions and Extracted Insights.}
After checking the generated questions and insights, we found pervasive quality deficiencies that likely led to low evaluation scores. First, some questions refer to column names that do not exist in the table, resulting in either missing insights or logical unsoundness. Second, several insights assert insufficient information, but the requisite data were present in the table. Furthermore, we observed examples of redundant or reduplicative insights.

\begin{figure*}[t!]
    \centering
    \includegraphics[width=\textwidth]{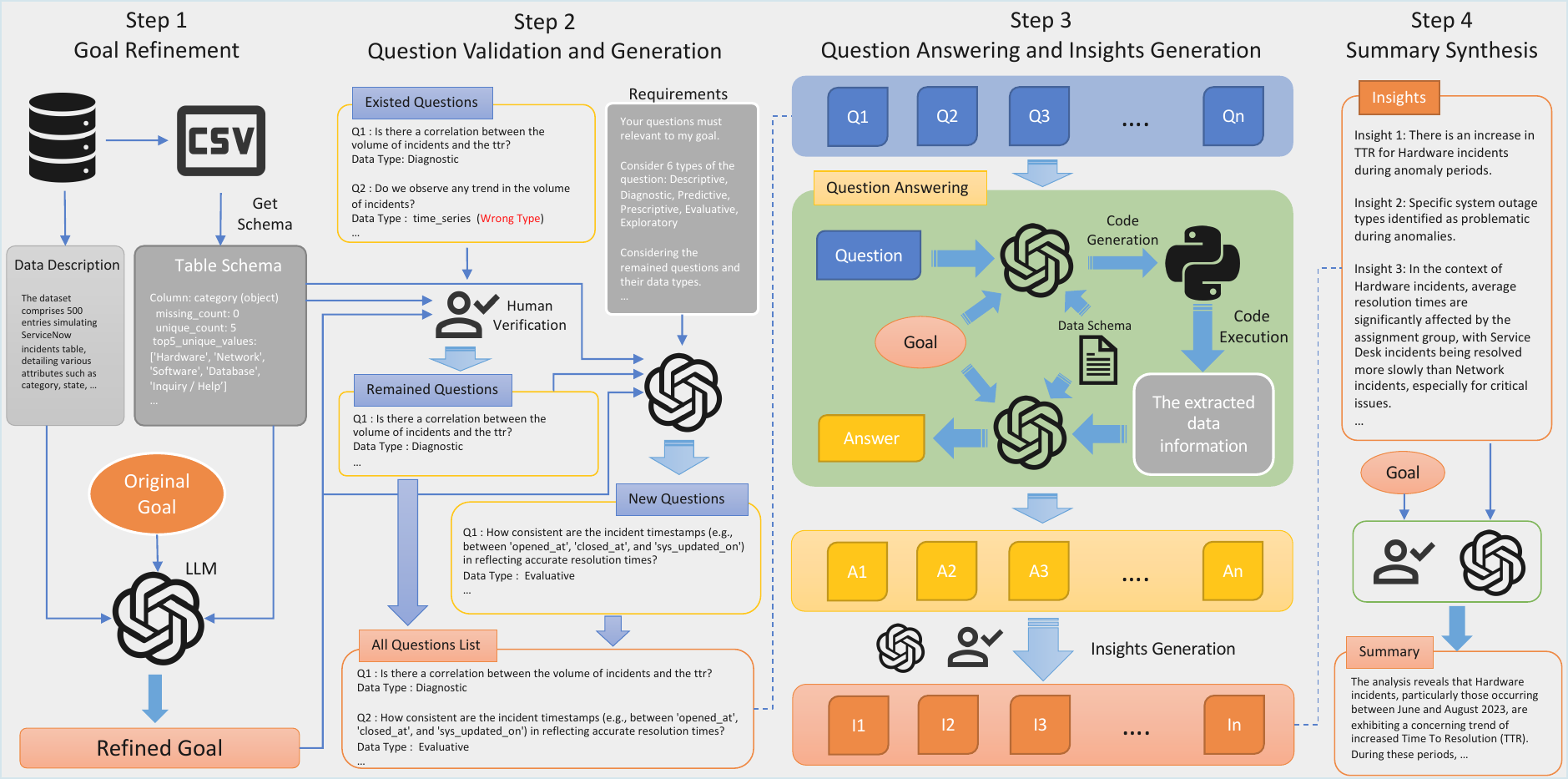}
    \caption{The dataset construction pipeline of InsightEval. This pipeline consists of 4 steps: 1) Refine the original goal through data description and schema. 2) Construct a new question list by verifying existing questions and generating new questions. 3) Extract key information by code generation and execution, then answer all the questions based on the refined goal, data schema, and extracted data, and finally generate insights. 4) Summarize all the insights by referring to the goal.}
    \label{fig:dataset_construction}
\end{figure*}

\paragraph{Observation 4: Insufficiently Comprehensive Evaluation Protocols.}
At present, evaluating insights is primarily based on automated text match metrics and G-Eval scoring. In InsightBench, it depends exclusively on LLAMA‑3‑Eval as the evaluator, thereby risking the model’s inherent biases. Moreover, the evaluation merely measures how many ground‑truth insights are matched by predicted insights while neglecting the generated erroneous insights. Lastly, it concentrates solely on discovering pre‑annotated insights and does not recognize or reward the discovery of novel insights.

\section{Dataset Construction Methodology and Evaluation Framework}

\subsection{Benchmark Requirements}
From the observation above, we summarize three critical requirements of a High-Quality Insight Benchmark as follows: 

\paragraph{R1. A Clearly Defined Task Goal.}
Each piece of data should be accompanied by a well-specified and unambiguous goal. This goal must explicitly state the intended analytical focus, including the relevant comparison metrics, dimensions of analysis, and quantitative evaluation criteria. 

\paragraph{R2. High-Quality Questions and Insights.}
Each question should be tightly aligned with the tabular data and the defined goal, avoiding any speculative or unfounded formulation. Moreover, the questions should be comprehensive, multidimensional, and reflective of diverse analytical perspectives, and the resulting insights should be meaningful, informative, and well-rounded.

\paragraph{R3. Multi-Perspective and Comprehensive Automatic Evaluation.}
The evaluation framework should incorporate multiple LLM evaluators to mitigate individual model biases and enhance scoring reliability. In addition to comprehensively evaluating the relevance and correctness of the generated insights, the evaluation should also assess the agent's ability to discover novel, valuable insights that are not present in the predefined ground-truth.

Based on the requirements outlined above, we constructed a new dataset for insight discovery. Each data point is accompanied by a clearly defined and unambiguous objective (\textbf{R1}), as well as meticulously curated, high‑quality questions and corresponding insights (\textbf{R2}). Building upon this dataset, we introduce a comprehensive evaluation framework for insight discovery that incorporates multi‑dimensional model scoring and explicitly accounts for a model’s capacity to autonomously discover novel insights (\textbf{R3}).

\subsection{Data Construction}
Insight Benchmark conventionally comprises four core components: a natural language objective (\textit{Goal}), a set of questions formulated to explore that objective (\textit{Questions}), the insights derived from the question‑answering process (\textit{Insights}), and a summary synthesized from all insights (\textit{Summary}).
Formally, each data instance can be shown as follows:
\begin{equation}
x_i = \bigl(T_i,\, G_i,\, Q_i,\, A_i,\, I_i,\, S_i\bigr),
\end{equation}
where $T_i$ denotes the tabular data (with schema), $G_i$ is the goal, $Q_i$ is the question set, $A_i$ and $I_i$ are the corresponding answers and insights, and $S_i$ is the final summary. In this paradigm, the goal and the tabular data serve as inputs, while the Insights and Summary constitute the ground‑truth outputs.

Our preliminary investigation in \textbf{R1} and \textbf{R2} established that an exemplary insight benchmark should satisfy three key criteria: 
(1) \textbf{Clear Objectives}: Each goal must specify the exact metrics and analytical dimensions to be compared; 
(2) \textbf{High‑Quality Questions}: Every question must be tightly related to both the Goal and the tabular data, and should be sufficiently broad and multi‑perspective; 
(3) \textbf{High‑Quality Insights}: Each insight should summarize from a rigorous analysis of the question‑answer pair, yielding substantive and data‑driven conclusions.

Most of the existing data science benchmarks require LLMs merely to answer isolated queries or questions (e.g., the InfiAgent‑DABench~\cite{hu2024infiagent}). Consequently, InsightBench~\cite{sahuinsightbench} is the closest match to our needs, offering 100 instances, 475 insights. However, we identified several structural and quality deficiencies. Accordingly, we have developed our dataset based on InsightBench, with the express aim of rectifying these issues and filling its gaps. The dataset is constructed via a hybrid pipeline combining manual inspection and LLM-assisted (o3-mini~\cite{o3mini2025openai}) generation. An overview of this pipeline is shown in Figure~\ref{fig:dataset_construction}. The detailed construction process of our dataset is described below:

\paragraph{Step 1: Goal Refinement.}
To ensure that each goal is precise and unambiguous, we implemented a meticulous validation pipeline combining LLM feedback with human review. First, given a table $T_i$, we extract its schema $\mathcal{S}_i$ (e.g., column names, data types, and basic statistics). Then, an LLM refines the initial goal based on the schema:
\begin{equation}
\tilde{G}_i = L_{\text{goal}}(\mathcal{S}_i).
\end{equation}
The refined goal is evaluated by human verification on three axes: alignment with the table, feasibility, and clarity of formulation. Only goals satisfying a predefined quality threshold are retained:
\begin{equation}
G_i =
\begin{cases}
\tilde{G}_i, & H_{\text{goal}}(\tilde{G}_i, \mathcal{S}_i) \ge \theta_{\text{goal}}, \\
\text{revision}, & \text{otherwise},
\end{cases}
\end{equation}
where $H_{\text{goal}}(\cdot)$ denotes human evaluation function.

\paragraph{Step 2: Question Generation and Validation.}
We first manually review the existing questions and retain those aligned with the refined goal, forming an initial set $Q_i^{(0)}$. To ensure comprehensive coverage, we categorize insight questions into six types: \emph{Descriptive}, \emph{Diagnostic}, \emph{Predictive}, \emph{Prescriptive}, \emph{Evaluative}, and \emph{Exploratory}. For each data instance, we require exactly ten questions and enforce coverage over all insight types:
\begin{equation}
\!\!\!|Q_i| = 10,
\forall \tau \in \mathcal{T},\ \exists q \in Q_i \ 
\text{s.t.}\ \mathrm{type}(q)=\tau
\end{equation}
where $\mathcal{T}$ denotes the set of six insight categories.

Additional questions are generated by an LLM conditioned on the table schema, the refined goal, and the retained questions:
\begin{equation}
Q_i^{\text{new}} = L_{\text{q}}(\mathcal{S}_i, G_i, Q_i^{(0)}).
\end{equation}
The final question set $Q_i$ is obtained by selecting valid questions from $Q_i^{(0)} \cup Q_i^{\text{new}}$ and passing them through human validation.

\paragraph{Step 3: Answering Questions and Generating Insights.}
For each question $q \in Q_i$, we generate answers and insights through a three-stage pipeline. First, LLM produces executable data analysis code:
\begin{equation}
C_{iq} = L_{\text{code}}(\mathcal{S}_i, G_i, q),
\end{equation}
which is executed locally on $T_i$ to obtain structured analytical outputs:
\begin{equation}
O_{iq} = \mathrm{Exec}(C_{iq}, T_i).
\end{equation}
Conditioned on these outputs, the LLM generates a concise factual answer:
\begin{equation}
a_{iq} = L_{\text{ans}}(O_{iq}, G_i, q),
\end{equation}
followed by an interpretive insight that contextualizes the answer with respect to the goal:
\begin{equation}
s_{iq} = L_{\text{ins}}(G_i, \mathcal{S}_i, q, a_{iq}).
\end{equation}
All generated answers and insights are manually verified for correctness and relevance. Invalid or misleading outputs are removed. To eliminate redundancy, we apply a semantic de-duplication operator over the insight set:
\begin{equation}
I_i \leftarrow \mathrm{Dedup}(I_i),
\end{equation}
and repeat question and insight generation until no new unique insights are introduced, i.e.,
\begin{equation}
|\mathrm{Dedup}(I_i^{(t)})| = |\mathrm{Dedup}(I_i^{(t-1)})|.
\end{equation}

\paragraph{Step 4: Summary Synthesis.}
Finally, given the refined goal, all validated questions, and their corresponding insights, the LLM synthesizes a comprehensive summary:
\begin{equation}
S_i = L_{\text{sum}}(G_i, Q_i, I_i).
\end{equation}
The summary is subsequently verified through human review to ensure conciseness, factual consistency, and actionable value.

\begin{table*}[th!]
\centering
\fontsize{8pt}{8pt}\selectfont 
\begin{tabular}{l|cccccc}
\toprule
\textbf{Dataset} & \textbf{Input} & \textbf{Output} & \textbf{Data Size} & \textbf{Construction Method} \\
\midrule
Spider 2.0~\cite{leispider} & Question & SQL Query & 632 Problems & Machine \& Human-Labeled \\
MatPlotBench~\cite{yang2024matplotagent} & Question+Table & Visual Image & 100 Cases & Machine \& Human-Labeled \\ 
InfiAgent-DABench~\cite{hu2024infiagent} & Question+Table & Answer & 603 Cases & Machine-Labeled \\
MedAgentsBench~\cite{tang2025medagentsbench} & Question & Answer & 862 Problems & Existed Dataset Combined \\
InsightBench~\cite{sahuinsightbench} & Goal+Table & Insights & 100 Cases (475 Insights) & Human-Labeled \\
\midrule
\textbf{InsightEval} & Goal+Table & Insights & 100 Cases (\textbf{1000} Insights) & \textbf{Machine \& Human-Refined} \\
\bottomrule
\end{tabular}
\caption{Comparison of InsightEval with other existing benchmarks.}
\label{Table:DatasetTable}
\end{table*}

\subsection{Evaluation Framework}

Evaluating the insight discovery capabilities of the agent on InsightEval requires comparing the agent-generated insights ($I$) with the annotated ground-truth insights ($GT$). To address the shortcomings identified in \textbf{Observation 4}, we propose a set of revised evaluation criteria, as articulated in \textbf{R3}, and design novel metrics accordingly. Our automated evaluation framework employs three principal measures in insights: recall, precision, $F1$, and novelty. With these measurements, we can comprehensively assess an agent’s ability in insights discovery. Furthermore, we also perform a dedicated evaluation of summary synthesis. Below, we detail each component of our methodology.

\paragraph{Insights Recall Evaluation.}
To assess if ground‑truth insights are discovered, we need to calculate the recall rate by adapting the iterative matching protocol. We count the scores between each ground‑truth insight ($gt \in GT$) and each agent-generated insight ($i \in I$). Then we record the highest‑scoring counterpart based on each ground-truth insight ($gt$) and calculate the expectation score ($E$) as the final output. The formula for recall evaluation is shown as in Equation~\ref{recall_equation}, with $\mathcal{S}$ representing the evaluator.
\begin{equation}
    \mathrm{Score_{recall}}
    \!\;=\!\;
    E_{gt \sim \mathrm{Unif}(GT)}
    \bigl[\max_{i \in I}\,\mathcal{S}(gt,i)\bigr]
    \label{recall_equation}
\end{equation}

\paragraph{Insights Precision Evaluation.}
Only focusing on the recall rate may overlook the possibility that agents generate irrelevant or unnecessary insights. To address this limitation, it is essential to further evaluate the accuracy of each generated insight to enhance the overall evaluation system. Similarly, we also enumerate the scores between the ground truth and the agent-generated insight. However, to calculate the precision rate, we need to record the highest score based on each agent-generated insight ($I$). The formula for precision evaluation is presented as in Equation~\ref{precision_equation}.
\begin{equation}
    \mathrm{Score_{precision}}
    \!\!\;=\!\!\;
    E_{i \sim \mathrm{Unif}(I)}
    \bigl[\max_{gt \in GT}\,\mathcal{S}(i,gt)\bigr]
    \label{precision_equation}
\end{equation}

\paragraph{Insights $F1$ Evaluation.}
To comprehensively evaluate the capability of insight discovery, we proposed a new measurement called Insight $F1$ Score. With the insight recall score and the insight precision score, we can calculate the insight $F1$ score through the formula in equation~\ref{F1_equation}.
\begin{equation}
    \mathrm{Score_{F1}}
    \!\!\;=\!\!\;
    \frac{2 * Score_{recall} * Score_{precision}}{Score_{recall}+Score_{precision}}
    \label{F1_equation}
\end{equation}

\paragraph{Insights Novelty Evaluation.}
Given the limitations of merely aligning with ground-truth insights, it is essential to evaluate the capacity of discovering novel insights. We identify insights with a G-Eval score exceeding 5 in the insight precision evaluation as correct, while the other insights are classified as incorrect and subjected to a secondary evaluation focused on innovation. During the evaluation, we utilize three distinct LLMs to mitigate bias. The insight can be labeled as a potential novel insight when at least two models judge it as correct. To get more accurate judgments, we provide LLMs with contextual information, including the goal, table schema, and historical insights, and use a Chain-of-Thought (CoT) reasoning framework. The formula for novelty evaluation is expressed as in Equation~\ref{novelty_equation}, where $\mathrm{LLM}_j(i)\in\{0,1\},\quad \delta\in\{0,1\}$, $j$ is the number of LLMs, $\mathbf{1}$ means indicator function, $M$ and $N$ indicate the number of correct and incorrect insights in precision evaluation, respectively.
\begin{equation}
    \mathrm{Score_{novelty}} = \frac{M + \delta \cdot C}{N + M}
    \label{novelty_equation}
\end{equation}
where $C$ is the count of novel samples:
\begin{equation}
    C = \sum_{i=1}^{N} \mathbf{1} \Bigl( \sum_{j=1}^{3} \mathrm{LLM}_j(i) \ge 2 \Bigr)
\end{equation}
When $\delta=1$, the formula will calculate the innovation score. For comparison, we set $\delta=0$ to obtain the original score during the evaluation. 

\paragraph{Summary Evaluation.}
For summaries, we perform a one‑to‑one comparison between each ground‑truth summary and its generated counterpart. Then we use evaluators to score each pair to derive an evaluation of summary quality.

\section{InsightEval: Statistic and Quality Analysis}

\subsection{Benchmark Statistic}
\begin{figure}
    \centering
    \includegraphics[width=\linewidth]{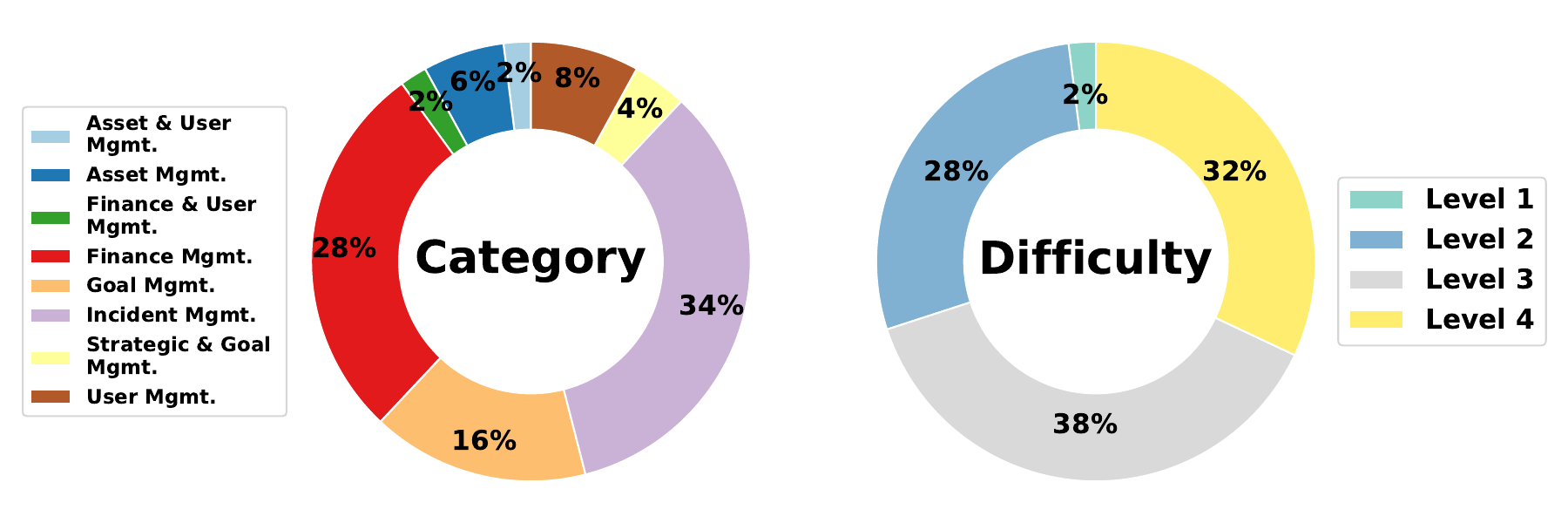}
    \caption{Data Statistics in InsightEval.}
    \label{fig:data_statistic}
\end{figure}

\begin{figure}
    \centering
    \includegraphics[width=0.9\linewidth]{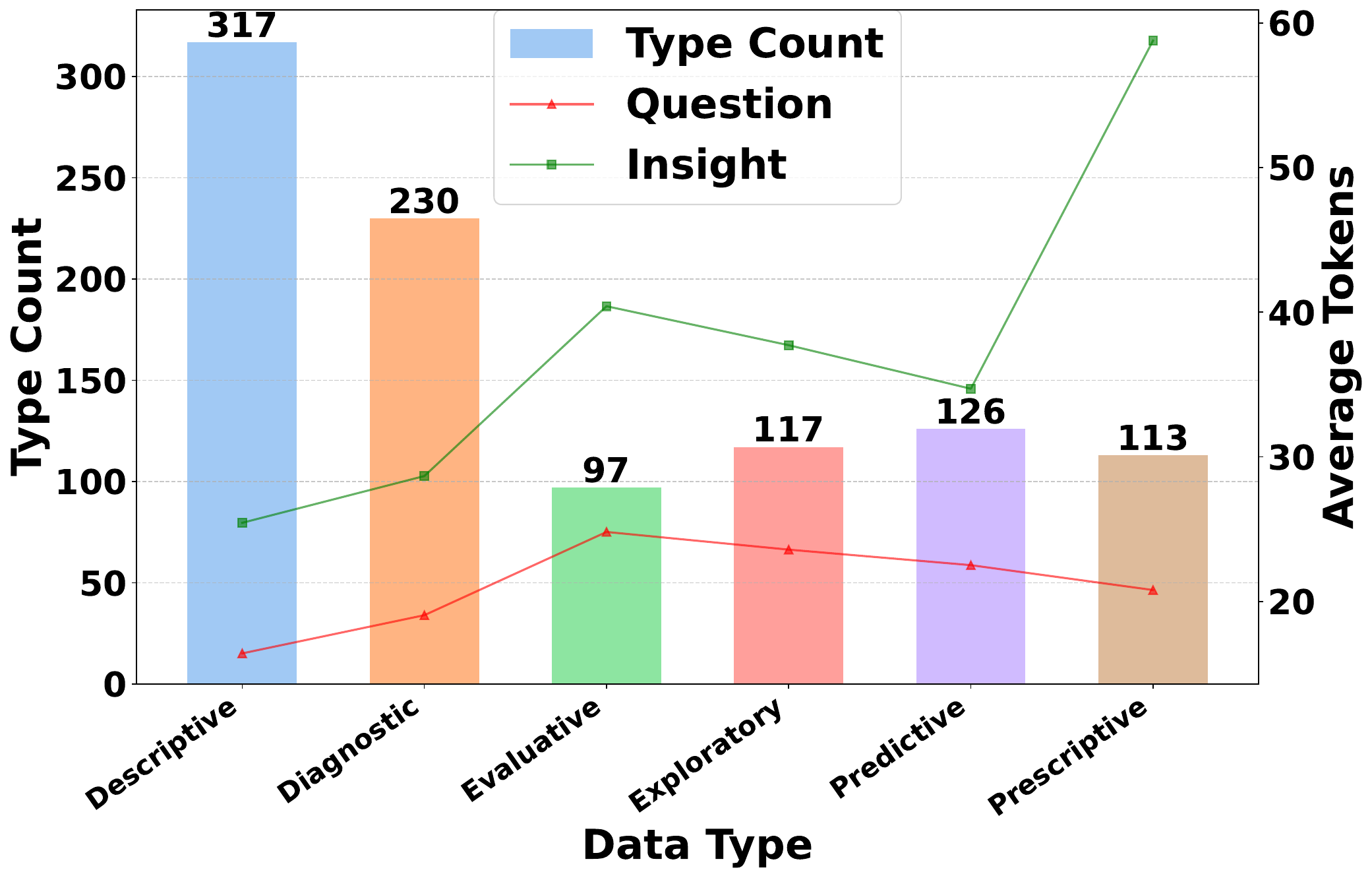}
    \caption{Data Type Distribution and Token Counts}
    \label{fig:typetoken}
\end{figure}

InsightEval comprises 100 instances, each with its corresponding CSV table. For each instance, we provide 10 individual insights and one overall summary. We adopt the difficulty and category established by InsightBench. Each data point is assigned one of four difficulty levels and also annotated with six commercial analytics scenarios, with distribution shown in Figure~\ref{fig:data_statistic}. In addition, we counted the number of each insight category and calculated the average tokens for the corresponding questions and insights, which are presented in Figure~\ref{fig:typetoken}. Moreover, compared to other well-regarded datasets, our InsightEval stands out for its large-scale and comprehensive insights, which are displayed in Table~\ref{Table:DatasetTable}.

\subsection{Data Quality Analysis}
\label{data_quality_analysis}
To further ensure data quality, we conducted an in-depth annotation combining evaluations by LLM (o3-mini \cite{o3mini2025openai}) and domain experts across three dimensions:

\begin{itemize}
\item \textbf{Correctness}: whether each question set strictly corresponds to the stated objective and the source table data without factual errors.
\item \textbf{Rationality}: whether each insight satisfies the goal’s requirements and is logically sound.
\item \textbf{Coherence}: whether insights are internally consistent and mutually compatible, including the overall summary's logical flow.
\end{itemize}

We annotated all instances by both LLM and human experts, computing the accuracy rate for each dimension. The results are reported in Table~\ref{data-quality}.

\begin{table}[h]
  \centering
  \fontsize{8.0pt}{8.0pt}\selectfont 
  \begin{tabular}{l|ccc}
    \toprule
    Method & Correctness & Rationality & Coherence \\
    \midrule
    LLM Evaluation & 0.905 & 0.890 & 0.90 \\
    Human Annotation & 0.935 & 0.920 & 0.95 \\
    \bottomrule
  \end{tabular}
  
  \caption{\label{data-quality}Results of data quality assessment.}
\end{table}

\begin{table}[h]
  \centering
  \fontsize{9pt}{9pt}\selectfont 
  \begin{tabular}{l|ccc}
    \toprule
    Type & TC Similarity & Self‑BLEU & Distinct‑2 \\
    \midrule
    Questions & 0.0583 & 0.1484 & 0.8677 \\
    Insights & 0.0528 & 0.0897 & 0.9157 \\
    \bottomrule
  \end{tabular}
  \caption{\label{redundant}Comparison of average TF‑IDF cosine (TC) similarity, Self‑BLEU, and Distinct‑2 diversity scores between Question and Insight in InsightEval.}
\end{table}

\begin{table*}[t!]
    \centering
    \fontsize{8.5pt}{8.5pt}\selectfont 
    \setlength{\tabcolsep}{3.4pt}
    \renewcommand{\arraystretch}{1.0}
    \begin{tabular}{l|cc|cc|cc|cc}
    \toprule
    \multirow{2}{*}{\textbf{Baselines}}  & \multicolumn{2}{c|}{\textbf{Insights Recall}} & \multicolumn{2}{c|}{\textbf{Insights Precision}} & \multicolumn{2}{c|}{\textbf{Insights $F_1$}} & \multicolumn{2}{c}{\textbf{Summary}}\\
     & ROUGE-1~~&~G-Eval & ROUGE-1~~&~G-Eval & ROUGE-1~~&~G-Eval & ROUGE-1~~&~G-Eval\\
    \midrule
    \rowcolor{lightgray}
    \multicolumn{9}{c}{\textbf{\textit{LLM-only}}} \\
    \midrule
    GPT-4o              & 0.2304 & 0.3389 & 0.2445 & 0.3506 & 0.2372 & 0.3447 & 0.2423 & 0.3282 \\
    DeepSeek-V3         & 0.2183 & 0.3402 & 0.2295 & 0.3554 & 0.2238 & 0.3476 & 0.2405 & 0.3332 \\
    Claude-3.7-Sonnet   & 0.2219 & 0.3265 & 0.2364 & 0.3492 & 0.2289 & 0.3375 & 0.2439 & 0.3250 \\
    \midrule
    \rowcolor{lightgray}
    \multicolumn{9}{c}{\textbf{\textit{Single-Agent}}} \\
    \midrule
    ReAct (GPT-4o)      & 0.2506 & 0.3977 & 0.2573 & 0.4069 & 0.2539 & 0.4022 & 0.2654 & 0.3913 \\
    CodeGen (GPT-4o)    & 0.2488 & 0.4289 & 0.2579 & 0.4412 & 0.2533 & 0.4350 & 0.2598 & 0.3991 \\
    \midrule
    \rowcolor{lightgray}
    \multicolumn{9}{c}{\textbf{\textit{Multi-Agents}}} \\
    \midrule
    DeepResearchAgent (GPT-4o)          & \underline{0.2993} & 0.5017 & \underline{0.3079} & 0.5198 & \underline{0.3035} & 0.5106 & \underline{0.3363} & 0.4279 \\
    Pandas Agent (GPT-4o)               & \textbf{0.3024} & 0.4973 & \textbf{0.3112} & 0.5133 & \textbf{0.3067} & 0.5052 & 0.3289 & 0.4021 \\
    Agent Poirot (GPT-4o)               & 0.2907 & \underline{0.5293} & 0.2945 & 0.5487 & 0.2926 & \underline{0.5388} & \textbf{0.3496} & 0.4334 \\
    Agent Poirot (Deepseek-V3)          & 0.2590 & 0.4984 & 0.2658 & 0.5453 & 0.2624 & 0.5208 & 0.3165 & \textbf{0.4772} \\
    Agent Poirot (Claude-3.7-Sonnet)    & 0.2623 & \textbf{0.5519} & 0.2673 & \textbf{0.6261} & 0.2648 & \textbf{0.5867} & 0.3178 & \underline{0.4746} \\
    \midrule
    Avg. & 0.2584 & 0.4411 & 0.2672 & 0.4657 & 0.2627 & 0.4529 & 0.2901 & 0.3992 \\

    \bottomrule
    \end{tabular}
    \caption{\label{insights_result}
    Comparative performance of various baselines at the Insights and Summary Level in InsightEval.}
\end{table*}

In addition, we checked each question and insight for redundancy in three metrics. First, we calculated cosine similarity over TF-IDF vector representations and averaged the resulting similarity scores. Second, we computed Self-BLEU of each sentence in questions and insights. Third, we measured Distinct-2, defined as the ratio of unique bi-grams to total bi-grams across all sentences. For TF-IDF cosine similarity and Self-BLEU scores, higher values indicate redundancy and more repetition. However, the Distinct-2 value that is closer to 1 reflects greater lexical diversity and lower redundancy. We present the redundancy statistics in Table~\ref{redundant}. Through this process, our dataset attains a high standard of reliability and scholarly validity.

\section{Experiments}

\subsection{Experimental Setup}

\paragraph{Baselines.} 
We evaluate various baselines on InsightEval using three prominent large language models currently in widespread use, such as GPT‑4o~\cite{openai2024gpt4ocard}, Deepseek‑V3~\cite{deepseekai2025deepseekv3technicalreport}, and Claude‑3.7‑Sonnet~\cite{anthropic2025claude37}. These LLMs are employed as a backbone across multiple agent-based frameworks, including ReAct~\cite{yao2023react}, CodeGen~\cite{majumder2025discoverybench}, DeepResearchAgent~\cite{zhang2025agentorchestraorchestratinghierarchicalmultiagent}, Pandas Agent~\cite{langchainpandas}, and Agent Poirot~\cite{sahuinsightbench}. Details of each agentic baseline are described in Appendix~\ref{app_baseline_details}.

\paragraph{Implementation Details.}
We configure each agent with a temperature of 0 to ensure determinism. In Agent Poirot, we run a total of 4 rounds, with 3 new questions generated in each round. Similarly, we also have another agent generate the same number of questions to ensure the rationality.

\paragraph{Evaluation Metrics.}
For insight recall, precision, and summary assessment, we employ two evaluators: the ROUGE‑1 \cite{lin2004rouge} and G-Eval \cite{liu2023g}. Specifically, the G-Eval score is the average score across GPT‑3.5-Turbo \cite{ye2023comprehensive} and Gemini 2.5 Pro \cite{comanici2025gemini}. Next, we use formula \ref{recall_equation} and \ref{precision_equation} to calculate the recall and precision scores. The final score in G-Eval should be normalized to facilitate a comparison with ROUGE-1. Then we calculate the insight $F1$ using the formula \ref{F1_equation}. For comparison, we sampled 30 data points and scored them by ten human experts. To measure insight novelty, we utilize the formula \ref{novelty_equation} to calculate the original novelty scores and new novelty scores.

\subsection{Experimental Results and Findings}

\paragraph{Insight $F1$ Score Provides a Better Reflection of Insight Capabilities.}
For insight $F1$ results shown in Table \ref{insights_result}, Pandas Agent achieved higher ROUGE‑1 scores, whereas Agent Poirot, based on the Claude 3.7 Sonnet, substantially outperformed others in G‑Eval. 

\begin{figure}
    \centering
    \includegraphics[width=\linewidth]{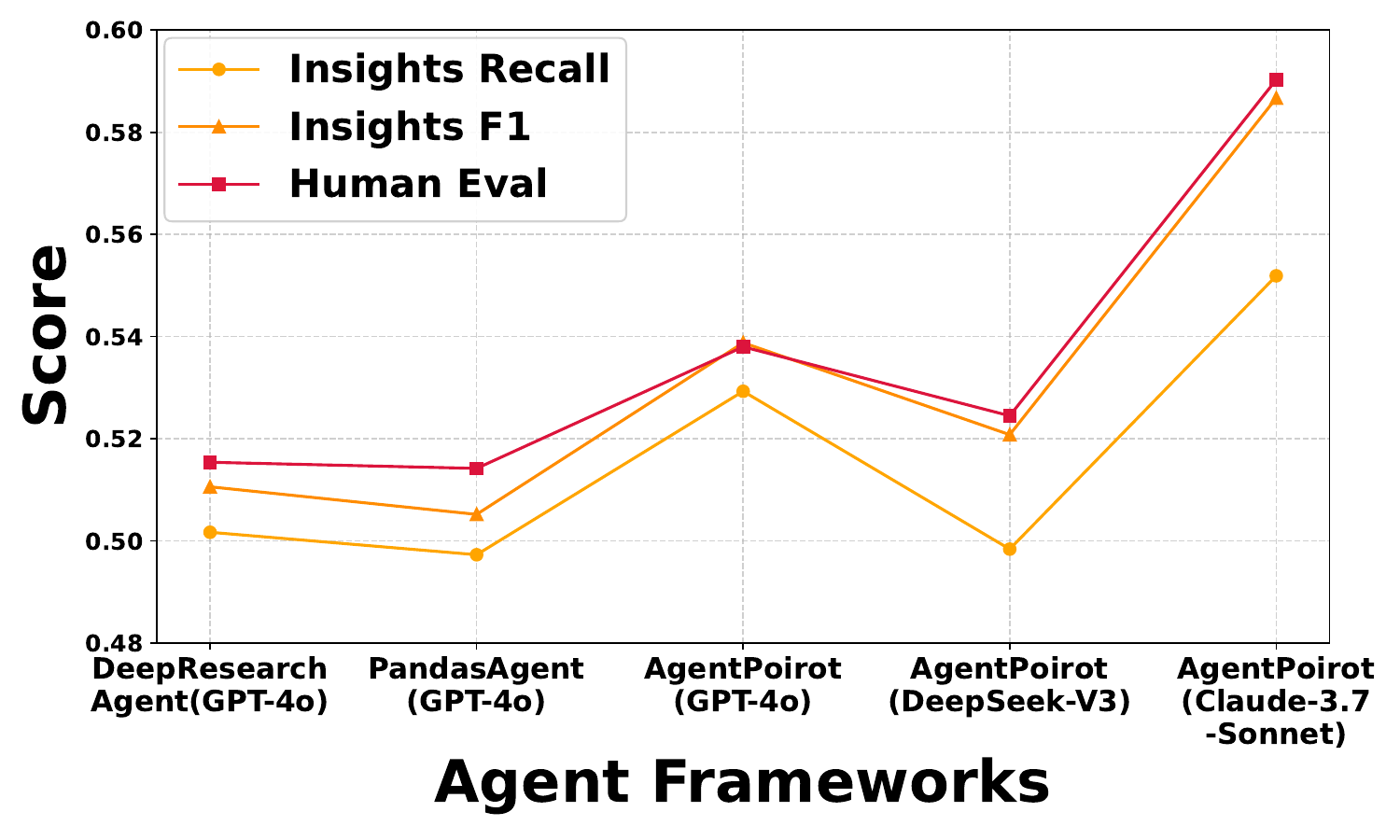}
    \caption{Comparison of G-Eval (Insight Recall \& Insight $F1$) and expert human-evaluation scores.}
    \label{fig:human_vs_f1}
\end{figure}

To further analyze the results, we conducted manual expert scoring, comparing it with Insight Recall (InsightBench Metric) and $F1$, which are shown in Figure \ref{fig:human_vs_f1}. Notably, Insights $F1$ Scores exceeded Insight Recall Scores across all agents, and were closer to the Human Evaluation Scores. This result underscores that Insights $F1$ Scores more effectively evaluate both the agent's insight discovery capability and its alignment with human judgment.

\paragraph{InsightEval Provides a Comprehensive Evaluation of the Agent's Performance.}
Our results reveal a range of challenges and key findings in insight discovery, as follows:

\textbf{Finding 1: Agents Exhibit Limited Breadth in Insight Exploration.}
In Table \ref{insights_result}, the Precision Scores of the insights outperform the recall scores, indicating the tendency that agents generate the most confidently correct insights while avoiding uncertain or exploratory output. This reduces random or spurious content, but results in substantial redundancy, which means highly scored insights may be correct and duplicated. Consequently, although agents demonstrate strong output quality, they come short in comprehensive exploration.

\textbf{Finding 2: Multi-Agent Systems Outperform Single-Model Baselines.}
As shown by the baseline comparisons in Table~\ref{insights_result}, multi-agent approaches achieve superior results on both the Insight and Summary evaluations. We attribute this advantage to the ability of multi-agent design to decompose the task pipeline and perform multiple rounds of targeted analysis and exploration, which enables more thorough reasoning and information synthesis. Concretely, according to G-Eval scores, Agent Poirot (Claude-3.7-Sonnet) achieves the highest Insight F1, while Agent Poirot (Deepseek-V3) achieves the highest Summary score. Moreover, the relative gain is slightly larger for Insight F1 than for the Summary metric.

\begin{figure}
    \centering
    \includegraphics[width=\linewidth]{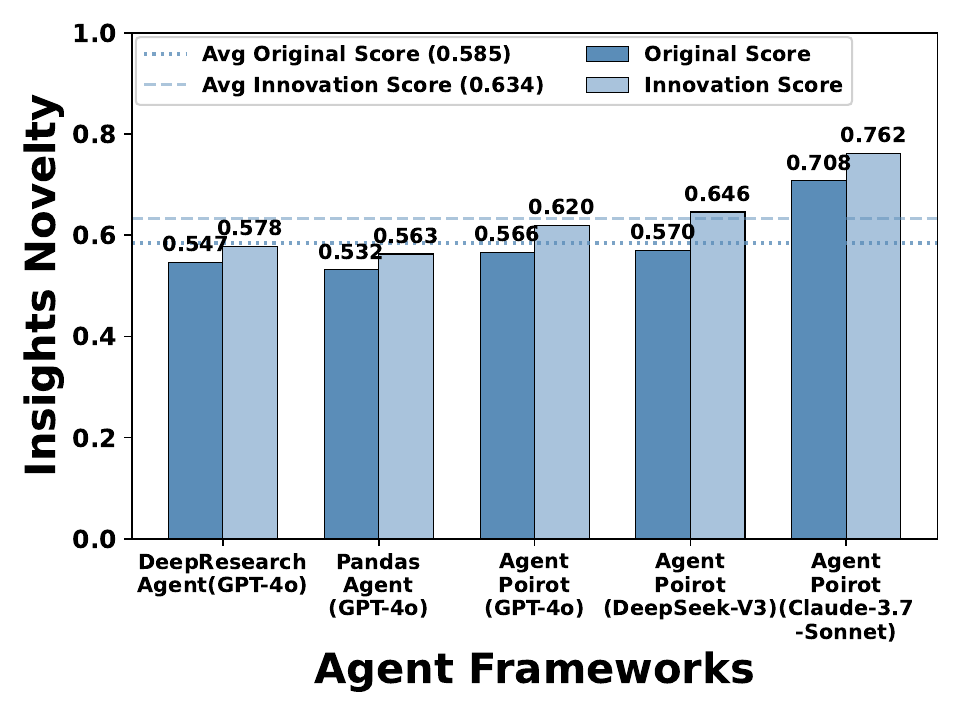}
    \caption{Performance of Novelty Evaluation.}
    \label{fig:novelty_result}
\end{figure}

\textbf{Finding 3: The Novelty of Agents Linked to the Capabilities of Backbone Models.}
We compared Original Novelty Scores with New Novelty Scores, as illustrated in Figure \ref{fig:novelty_result}. We observed that all agents achieved measurable improvements in novelty. In particular, the agent built on the Claude‑3.7‑Sonnet achieved the highest New Novelty Score of 76.2\%. We attribute this performance to its superior code‑generation abilities, which likely enable it to answer questions more effectively and derive deeper insights. By calculating the improvement ratio of the new novelty score over the original novelty score, we discovered that the Deepseek‑V3-based agent achieves the largest rate among all agents, with 13.3\%. This finding suggests that agents with lower precision may compensate by producing more creative outputs, thereby attaining higher novelty scores.

\section{Conclusion and Future Work}

We present InsightEval, a novel benchmark for rigorously assessing agent's insight discovery capabilities. Our contributions include proposing a comprehensive and high-quality dataset, developing an automated evaluation framework that spans recall, precision, and novelty, and conducting a systematic evaluation of state-of-the-art agent frameworks. Our findings uncover key challenges in automated insight discovery and offer valuable guidance for future research. In future work, we will design and evaluate state-of-the-art multi-agent systems to significantly enhance insight discovery, enabling paradigm-shifting advances in the domain.

\section*{Limitations}

This work introduces InsightEval as an expert-curated benchmark for table-driven insight generation, defining a structured taxonomy, evaluation criteria, and a standardized assessment protocol validated across multiple agents and language models. However, several inherent limitations remain difficult to eliminate. i) The dataset scale and domain coverage are constrained by annotation cost and design choices, and therefore may not fully reflect the diversity of real-world settings. To address this limitation, future work will explore controlled data augmentation and synthesis to enrich underrepresented scenarios, employ active and diversity-aware sampling strategies to prioritize informative and low-coverage cases, and expand domain coverage through collaborations with domain experts and community-driven data contributions. ii) The nature of an insight is intrinsically subjective, as judgments of value, actionability, and novelty vary across users and contexts. The pure table-input setting also imposes a natural information ceiling, since many meaningful insights require external or contextual knowledge not present in the data. Ground-truth annotations are necessarily incomplete and should be viewed as representative rather than exhaustive, and novelty assessments remain time- and context-dependent as reference knowledge evolves. As a result, findings should be interpreted as relative and conditional rather than absolute. In future work, we aim to expand evaluation contexts and longitudinal settings to understand better how model capabilities evolve and adapt across domains and time.

\bibliography{custom}

@inproceedings{yin2023natural,
    title = "Natural Language to Code Generation in Interactive Data Science Notebooks",
    author = "Yin, Pengcheng  and
      Li, Wen-Ding  and
      Xiao, Kefan  and
      Rao, Abhishek  and
      Wen, Yeming  and
      Shi, Kensen  and
      Howland, Joshua  and
      Bailey, Paige  and
      Catasta, Michele  and
      Michalewski, Henryk  and
      Polozov, Oleksandr  and
      Sutton, Charles",
    editor = "Rogers, Anna  and
      Boyd-Graber, Jordan  and
      Okazaki, Naoaki",
    booktitle = "Proceedings of the 61st Annual Meeting of the Association for Computational Linguistics (Volume 1: Long Papers)",
    month = jul,
    year = "2023",
    address = "Toronto, Canada",
    publisher = "Association for Computational Linguistics",
    url = "https://aclanthology.org/2023.acl-long.9/",
    doi = "10.18653/v1/2023.acl-long.9",
    pages = "126--173"
}

@article{lu2026multivis,
  author    = {Lu, Jinwei and Song, Yuanfeng and Zhang, Chen and Wong, Raymond Chi-Wing},
  title     = {MultiVis-Agent: A Multi-Agent Framework with Logic Rules for Reliable and Comprehensive Cross-Modal Data Visualization},
  journal   = {Proc. ACM Manag. Data},
  volume    = {4},
  number    = {1},
  articleno = {56},
  year      = {2026},
  month     = {feb},
  publisher = {ACM},
  doi       = {10.1145/3786670}
}

@inproceedings{lin2004rouge,
  title={Rouge: A package for automatic evaluation of summaries},
  author={Lin, Chin-Yew},
  booktitle={Text summarization branches out},
  pages={74--81},
  year={2004}
}

@inproceedings{liu2023g,
    title = "{G}-Eval: {NLG} Evaluation using Gpt-4 with Better Human Alignment",
    author = "Liu, Yang  and
      Iter, Dan  and
      Xu, Yichong  and
      Wang, Shuohang  and
      Xu, Ruochen  and
      Zhu, Chenguang",
    editor = "Bouamor, Houda  and
      Pino, Juan  and
      Bali, Kalika",
    booktitle = "Proceedings of the 2023 Conference on Empirical Methods in Natural Language Processing",
    month = dec,
    year = "2023",
    address = "Singapore",
    publisher = "Association for Computational Linguistics",
    url = "https://aclanthology.org/2023.emnlp-main.153/",
    doi = "10.18653/v1/2023.emnlp-main.153",
    pages = "2511--2522"
}

@article{anthropic2025claude37,
  title={Claude 3.7 Sonnet and Claude Code},
  author={Anthropic, AI},
  url={https://www.anthropic.com/news/claude-3-7-sonnet},
  year={2025}
}

@article{comanici2025gemini,
  title={Gemini 2.5: Pushing the frontier with advanced reasoning, multimodality, long context, and next generation agentic capabilities},
  author={Comanici, Gheorghe and Bieber, Eric and Schaekermann, Mike and Pasupat, Ice and Sachdeva, Noveen and Dhillon, Inderjit and Blistein, Marcel and Ram, Ori and Zhang, Dan and Rosen, Evan and others},
  journal={arXiv preprint arXiv:2507.06261},
  year={2025}
}

@misc{deepseekai2025deepseekv3technicalreport,
      title={DeepSeek-V3 Technical Report}, 
      author={DeepSeek-AI and Aixin Liu and Bei Feng and Bing Xue and Bingxuan Wang and Bochao Wu and Chengda Lu and Chenggang Zhao and Chengqi Deng and Chenyu Zhang and Chong Ruan and Damai Dai and Daya Guo and Dejian Yang and Deli Chen and Dongjie Ji and others},
      year={2025},
      eprint={2412.19437},
      archivePrefix={arXiv},
      primaryClass={cs.CL},
      url={https://arxiv.org/abs/2412.19437}, 
}

@misc{openai2024gpt4ocard,
      title={GPT-4o System Card}, 
      author={OpenAI},
      year={2024},
      eprint={2410.21276},
      archivePrefix={arXiv},
      primaryClass={cs.CL},
      url={https://arxiv.org/abs/2410.21276}, 
}

@article{ye2023comprehensive,
  title={A comprehensive capability analysis of gpt-3 and gpt-3.5 series models},
  author={Ye, Junjie and Chen, Xuanting and Xu, Nuo and Zu, Can and Shao, Zekai and Liu, Shichun and Cui, Yuhan and Zhou, Zeyang and Gong, Chao and Shen, Yang and others},
  journal={arXiv preprint arXiv:2303.10420},
  year={2023}
}

@article{o3mini2025openai,
  title={OpenAI o3-mini System Card},
  author={OpenAI},
  year={2025},
  url={https://cdn.openai.com/o3-mini-system-card-feb10.pdf}
}

@article{langchainpandas,
    author = {LangChain},
    title = {Pandas dataframe},
    journal = {},
    year = {2024},
    URL = {https://python.langchain.com/v0.2/docs/integrations/tools/pandas/}
}

@inproceedings{sahuinsightbench,
  title={InsightBench: Evaluating Business Analytics Agents Through Multi-Step Insight Generation},
  author={Sahu, Gaurav and Puri, Abhay and Rodriguez, Juan A and Abaskohi, Amirhossein and Chegini, Mohammad and Drouin, Alexandre and Taslakian, Perouz and Zantedeschi, Valentina and Lacoste, Alexandre and Vazquez, David and others},
year={2025},
  booktitle={The Thirteenth International Conference on Learning Representations}
}

@inproceedings{hu2024infiagent,
  title={InfiAgent-DABench: evaluating agents on data analysis tasks},
  author={Hu, Xueyu and Zhao, Ziyu and Wei, Shuang and Chai, Ziwei and Ma, Qianli and Wang, Guoyin and Wang, Xuwu and Su, Jing and Xu, Jingjing and Zhu, Ming and others},
  booktitle={Proceedings of the 41st International Conference on Machine Learning},
  pages={19544--19572},
  year={2024}
}

@inproceedings{lai2023ds,
  title={DS-1000: A natural and reliable benchmark for data science code generation},
  author={Lai, Yuhang and Li, Chengxi and Wang, Yiming and Zhang, Tianyi and Zhong, Ruiqi and Zettlemoyer, Luke and Yih, Wen-tau and Fried, Daniel and Wang, Sida and Yu, Tao},
  booktitle={International Conference on Machine Learning},
  pages={18319--18345},
  year={2023},
  organization={PMLR}
}

@article{chandel2022training,
  title={Training and evaluating a jupyter notebook data science assistant},
  author={Chandel, Shubham and Clement, Colin B and Serrato, Guillermo and Sundaresan, Neel},
  journal={arXiv preprint arXiv:2201.12901},
  year={2022}
}

@inproceedings{qin2025multitend,
  title={MultiTEND: A Multilingual Benchmark for Natural Language to NoSQL Query Translation},
  author={Qin, Zhiqian and Song, Yuanfeng and Lu, Jinwei and Song, Yuanwei and Li, Shuaimin and Zhang, Chen Jason},
  booktitle={Findings of the Association for Computational Linguistics: ACL 2025},
  pages={24632--24657},
  year={2025}
}

@article{lu2025bridging,
  title={Bridging the gap: Enabling natural language queries for nosql databases through text-to-nosql translation},
  author={Lu, Jinwei and Song, Yuanfeng and Qin, Zhiqian and Zhang, Haodi and Zhang, Chen and Wong, Raymond Chi-Wing},
  journal={arXiv preprint arXiv:2502.11201},
  year={2025}
}

@inproceedings{leispider,
  title={Spider 2.0: Evaluating Language Models on Real-World Enterprise Text-to-SQL Workflows},
  author={Lei, Fangyu and Chen, Jixuan and Ye, Yuxiao and Cao, Ruisheng and Shin, Dongchan and SU, Hongjin and SUO, ZHAOQING and Gao, Hongcheng and Hu, Wenjing and Yin, Pengcheng and others},
  booktitle={The Thirteenth International Conference on Learning Representations},
  year={2024}
}

@article{lee2022ehrsql,
  title={Ehrsql: A practical text-to-sql benchmark for electronic health records},
  author={Lee, Gyubok and Hwang, Hyeonji and Bae, Seongsu and Kwon, Yeonsu and Shin, Woncheol and Yang, Seongjun and Seo, Minjoon and Kim, Jong-Yeup and Choi, Edward},
  journal={Advances in Neural Information Processing Systems},
  volume={35},
  pages={15589--15601},
  year={2022}
}

@inproceedings{liu2025nl2sql,
  title={Nl2sql-bugs: A benchmark for detecting semantic errors in NL2SQL translation},
  author={Liu, Xinyu and Shen, Shuyu and Li, Boyan and Tang, Nan and Luo, Yuyu},
  booktitle={Proceedings of the 31th ACM SIGKDD Conference on Knowledge Discovery and Data Mining},
  year={2025}
}

@article{chen2024viseval,
  title={Viseval: A benchmark for data visualization in the era of large language models},
  author={Chen, Nan and Zhang, Yuge and Xu, Jiahang and Ren, Kan and Yang, Yuqing},
  journal={IEEE Transactions on Visualization and Computer Graphics},
  year={2024},
  publisher={IEEE}
}

@inproceedings{dong2025practiq,
  title={PRACTIQ: A Practical Conversational Text-to-SQL dataset with Ambiguous and Unanswerable Queries},
  author={Dong, Mingwen and Kumar, Nischal Ashok and Hu, Yiqun and Chauhan, Anuj and Hang, Chung-Wei and Chang, Shuaichen and Pan, Lin and Lan, Wuwei and Zhu, Henghui and Jiang, Jiarong and others},
  booktitle={Proceedings of the 2025 Conference of the Nations of the Americas Chapter of the Association for Computational Linguistics: Human Language Technologies (Volume 1: Long Papers)},
  pages={255--273},
  year={2025}
}

@article{xu2025dagent,
  title={Dagent: A relational database-driven data analysis report generation agent},
  author={Xu, Wenyi and Mao, Yuren and Zhang, Xiaolu and Zhang, Chao and Dong, Xuemei and Zhang, Mengfei and Gao, Yunjun},
  journal={arXiv preprint arXiv:2503.13269},
  year={2025}
}

@inproceedings{perez2025llm,
  title={An LLM-Based Approach for Insight Generation in Data Analysis},
  author={P{\'e}rez, Alberto S{\'a}nchez and Boukhary, Alaa and Papotti, Paolo and Lozano, Luis Castej{\'o}n and Elwood, Adam},
  booktitle={Proceedings of the 2025 Conference of the Nations of the Americas Chapter of the Association for Computational Linguistics: Human Language Technologies (Volume 1: Long Papers)},
  pages={562--582},
  year={2025}
}

@inproceedings{yang2024matplotagent,
    title = "{M}at{P}lot{A}gent: Method and Evaluation for {LLM}-Based Agentic Scientific Data Visualization",
    author = "Yang, Zhiyu  and
      Zhou, Zihan  and
      Wang, Shuo  and
      Cong, Xin  and
      Han, Xu  and
      Yan, Yukun  and
      Liu, Zhenghao  and
      Tan, Zhixing  and
      Liu, Pengyuan  and
      Yu, Dong  and
      Liu, Zhiyuan  and
      Shi, Xiaodong  and
      Sun, Maosong",
    editor = "Ku, Lun-Wei  and
      Martins, Andre  and
      Srikumar, Vivek",
    booktitle = "Findings of the Association for Computational Linguistics: ACL 2024",
    month = aug,
    year = "2024",
    address = "Bangkok, Thailand",
    publisher = "Association for Computational Linguistics",
    url = "https://aclanthology.org/2024.findings-acl.701/",
    doi = "10.18653/v1/2024.findings-acl.701",
    pages = "11789--11804"
}

@inproceedings{ouyang2025nvagent,
    title = "nv{A}gent: Automated Data Visualization from Natural Language via Collaborative Agent Workflow",
    author = "Ouyang, Geliang  and
      Chen, Jingyao  and
      Nie, Zhihe  and
      Gui, Yi  and
      Wan, Yao  and
      Zhang, Hongyu  and
      Chen, Dongping",
    editor = "Che, Wanxiang  and
      Nabende, Joyce  and
      Shutova, Ekaterina  and
      Pilehvar, Mohammad Taher",
    booktitle = "Proceedings of the 63rd Annual Meeting of the Association for Computational Linguistics (Volume 1: Long Papers)",
    month = jul,
    year = "2025",
    url = "https://aclanthology.org/2025.acl-long.960/",
    pages = "19534--19567"
}

@inproceedings{ma2023insightpilot,
    title = "{I}nsight{P}ilot: An {LLM}-Empowered Automated Data Exploration System",
    author = "Ma, Pingchuan  and
      Ding, Rui  and
      Wang, Shuai  and
      Han, Shi  and
      Zhang, Dongmei",
    editor = "Feng, Yansong  and
      Lefever, Els",
    booktitle = "Proceedings of the 2023 Conference on Empirical Methods in Natural Language Processing: System Demonstrations",
    month = dec,
    year = "2023",
    address = "Singapore",
    publisher = "Association for Computational Linguistics",
    url = "https://aclanthology.org/2023.emnlp-demo.31/",
    doi = "10.18653/v1/2023.emnlp-demo.31",
    pages = "346--352"
}

@article{weng2025insightlens,
  title={InsightLens: Augmenting LLM-Powered Data Analysis with Interactive Insight Management and Navigation},
  author={Weng, Luoxuan and Wang, Xingbo and Lu, Junyu and Feng, Yingchaojie and Liu, Yihan and Feng, Haozhe and Huang, Danqing and Chen, Wei},
  journal={IEEE Transactions on Visualization and Computer Graphics},
  year={2025},
  publisher={IEEE}
}

@inproceedings{wu2024autogen,
  title={Autogen: Enabling next-gen LLM applications via multi-agent conversations},
  author={Wu, Qingyun and Bansal, Gagan and Zhang, Jieyu and Wu, Yiran and Li, Beibin and Zhu, Erkang and Jiang, Li and Zhang, Xiaoyun and Zhang, Shaokun and Liu, Jiale and others},
  booktitle={First Conference on Language Modeling},
  year={2024}
}

@inproceedings{yao2023react,
  title={React: Synergizing reasoning and acting in language models},
  author={Yao, Shunyu and Zhao, Jeffrey and Yu, Dian and Du, Nan and Shafran, Izhak and Narasimhan, Karthik and Cao, Yuan},
  booktitle={International Conference on Learning Representations (ICLR)},
  year={2023}
}

@article{tang2025medagentsbench,
  title={Medagentsbench: Benchmarking thinking models and agent frameworks for complex medical reasoning},
  author={Tang, Xiangru and Shao, Daniel and Sohn, Jiwoong and Chen, Jiapeng and Zhang, Jiayi and Xiang, Jinyu and Wu, Fang and Zhao, Yilun and Wu, Chenglin and Shi, Wenqi and others},
  journal={arXiv preprint arXiv:2503.07459},
  year={2025}
}

@inproceedings{
majumder2025discoverybench,
title={DiscoveryBench: Towards Data-Driven Discovery with Large Language Models},
author={Bodhisattwa Prasad Majumder and Harshit Surana and Dhruv Agarwal and Bhavana Dalvi Mishra and Abhijeetsingh Meena and Aryan Prakhar and Tirth Vora and Tushar Khot and Ashish Sabharwal and Peter Clark},
booktitle={The Thirteenth International Conference on Learning Representations},
year={2025},
url={https://openreview.net/forum?id=vyflgpwfJW}
}

@misc{zhang2025agentorchestraorchestratinghierarchicalmultiagent,
      title={AgentOrchestra: Orchestrating Hierarchical Multi-Agent Intelligence with the Tool-Environment-Agent(TEA) Protocol}, 
      author={Wentao Zhang and Liang Zeng and Yuzhen Xiao and Yongcong Li and Ce Cui and Yilei Zhao and Rui Hu and Yang Liu and Yahui Zhou and Bo An},
      year={2025},
      eprint={2506.12508},
      archivePrefix={arXiv},
      primaryClass={cs.AI},
      url={https://arxiv.org/abs/2506.12508}, 
}

\clearpage
\onecolumn
\newpage
\twocolumn
\appendix

\section{Details of InsightEval Benchmark}
\subsection{Category and Difficulty}
Referring to the categories and difficulty settings in InsightBench~\cite{sahuinsightbench}, InsightEval contains 8 business analytics categories and 4 difficulty levels. We describe the details of each category as follows:
\begin{itemize}
    \item \textbf{Incident Management}: Tracks and analyzes operational or safety incidents to enable rapid response and root-cause investigation.
    \item \textbf{Asset Management}: Manages IT hardware lifecycle—procurement, deployment, maintenance—ensuring inventory visibility and optimal utilization.
    \item \textbf{User Management}: Maintains user profiles, roles, departments, and login status to enforce access control and support audit trails.
    \item \textbf{Finance Management}: Audits expense records to reveal spending patterns, optimize budget allocation, and drive cost-saving decisions.
    \item \textbf{Goal Management}: Monitors departmental or project objectives—planning, progress, and completion rate—to assess performance and alignment.
    \item \textbf{Asset \& User Management}: Correlates hardware assignments with user data to optimize resource distribution and usage efficiency.
    \item \textbf{Finance \& User Management}: Links expense transactions with individual or team activity to uncover cost behaviors and usage trends.
    \item \textbf{Strategy \& Goal Management}: Integrates strategic plans with goal-tracking data to evaluate execution effectiveness and organizational alignment.
\end{itemize}

\begin{figure}[t!]
    \centering
    \includegraphics[width=\linewidth]{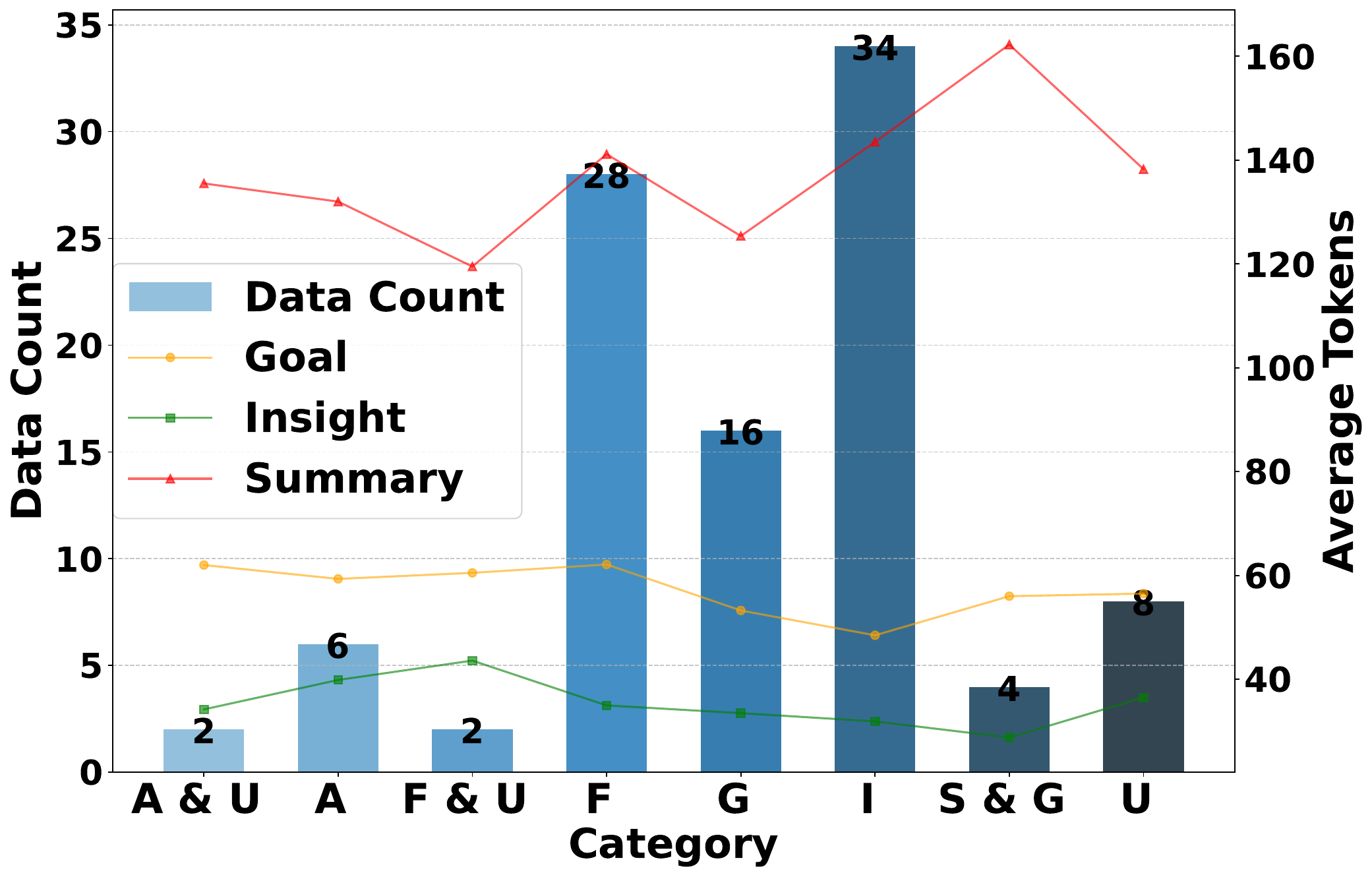}
    \caption{Category distributions with token count in different data items. A, U, F, G, I, S separately stand for Asset, User, Finance, Goal, Incident, Strategy.}
    \label{fig:cate_token}
\end{figure}

\begin{figure}[t!]
    \centering
    \includegraphics[width=\linewidth]{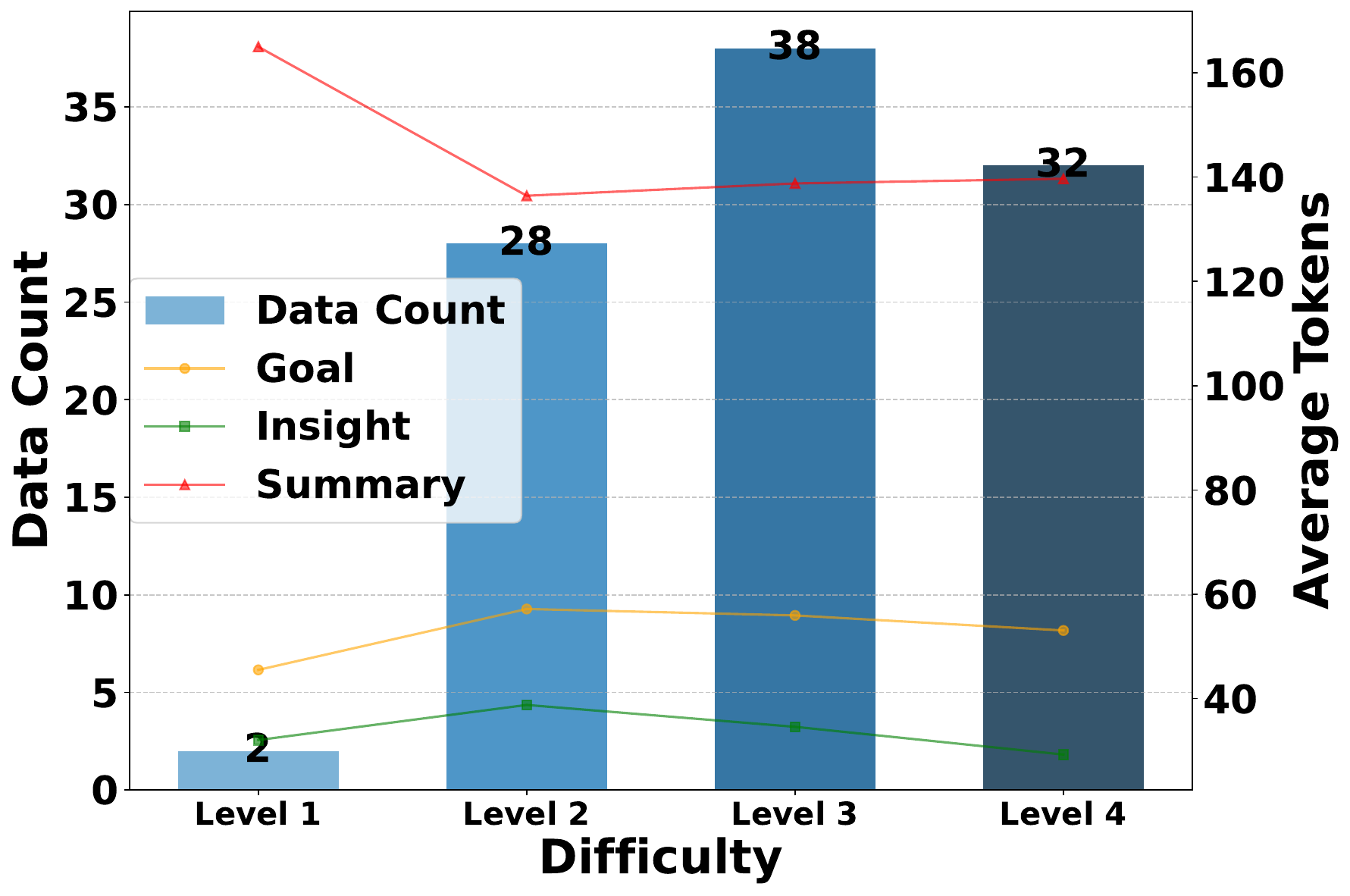}
    \caption{Difficulty distributions with token count in different data items.}
    \label{fig:diff_token}
\end{figure}

\subsection{Insight Types}
In this paper, we aim to provide a more comprehensive interpretation of data insights by extending the four insight categories originally defined in InsightBench~\cite{sahuinsightbench} with two additional types. A detailed description of each insight category is provided below:

\begin{itemize}
    \item \textbf{Descriptive}: Summarize what happened. This type of analysis describes past situations by aggregating and visualizing historical data (for example, generating a chart of monthly investment portfolio returns). 
    \item \textbf{Diagnostic}: Explain why it happened. Identifies correlations, patterns, and root causes to explain observed trends or results (for example, segmenting losses by asset category to identify key drivers).
    \item \textbf{Predictive}: Forecast what is likely to happen. Based on historical trends, it uses statistical models to predict future outcomes (for example, estimating the risk of default in the next quarter using credit scores).
    \item \textbf{Prescriptive}: Recommend specific actions to take. It suggests actionable strategies, such as optimization or risk mitigation, to achieve desired objectives (for example, advising portfolio adjustments to reduce predicted volatility).
    \item \textbf{Evaluative}: Assess the quality and reliability of the data and analysis. This involves evaluating the completeness, accuracy, and robustness of data and analytical methods (for example, verifying data integrity in critical fields or conducting back-tests to validate the accuracy of a risk model).
    \item \textbf{Exploratory}: Discover hidden patterns or anomalies. Without predefined hypotheses, it explores data freely using visualization and statistical techniques to uncover unknown relationships, structures, clusters, or outliers (for example, applying clustering methods to identify unexpected customer segments or detecting abnormal transactions).
\end{itemize}

\subsection{Tokens Count Analysis}
In addition, we have statistics on the distribution of counts across categories, difficulty levels, and insight types. We also computed the average token length of components of the Goal, Insight, and Summary components for the corresponding data. The token length statistics for categories and difficulty levels are illustrated in Figures \ref{fig:cate_token} and \ref{fig:diff_token}, while the statistics for insight categories are illustrated in Figure 4 of the main text.

As observed in the category and difficulty-level statistics, the Summary section, which synthesizes all insights, exhibits the longest average token length, whereas Insights show the shortest. When examining different insight categories, the token lengths for Questions remain relatively consistent, while Insights in the Prescriptive category demonstrate the highest token count. This is because the Prescriptive insight type typically requires the provision of extensive optimization measures, recommendations, and strategic suggestions.

\section{Details of Human Review}
We conducted an extensive manual review and annotation in this paper. First, during error analysis of existing datasets, we systematically examined each dataset’s goals, questions, and insights. Next, in constructing the InsightEval dataset, we inspected and annotated preexisting issues and then manually validated the outputs of each generation step. For quality control, a panel of domain experts evaluated the dataset across multiple criteria. Finally, to assess the validity and reliability of our proposed insight metrics, we solicited expert ratings and compared these annotations against our automated measures for rigorous analysis.

\section{InsightEval vs. InsightBench}
\begin{figure}[t!]
    \centering
    \includegraphics[width=\linewidth]{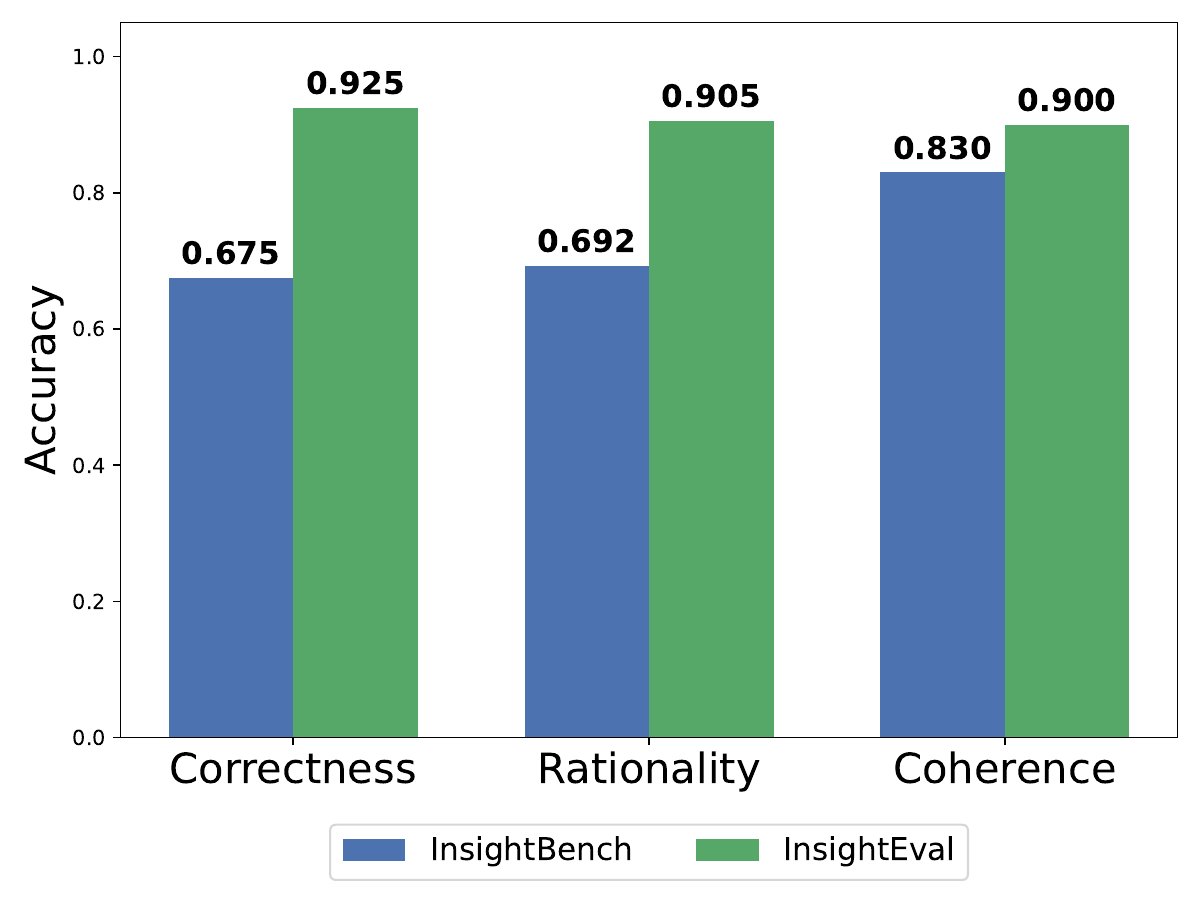}
    \caption{Comparative Benchmark Quality Assessment via LLM Evaluation}
    \label{fig:bench_comp_llm}
\end{figure}

\begin{figure}[t!]
    \centering
    \includegraphics[width=\linewidth]{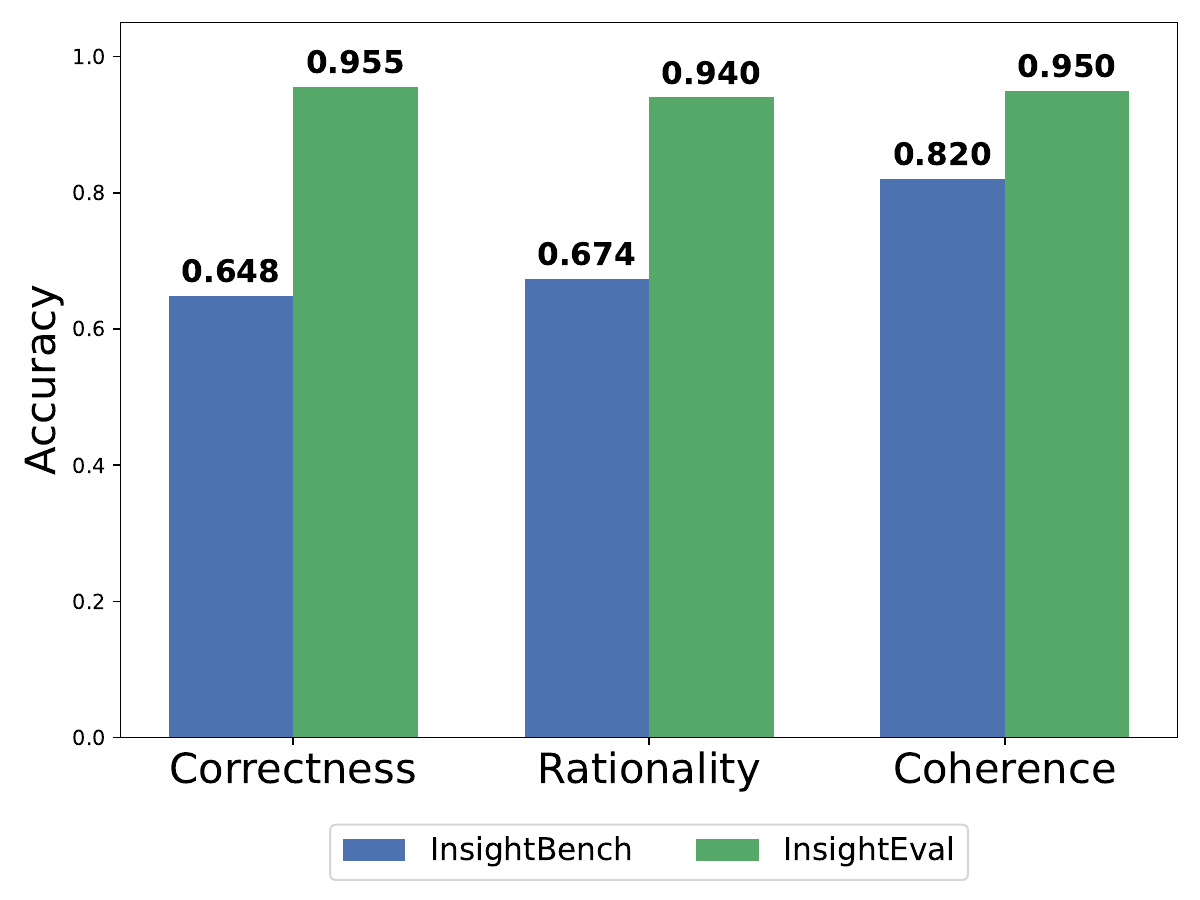}
    \caption{Comparative Benchmark Quality Assessment via Human Evaluation}
    \label{fig:bench_comp_human}
\end{figure}

To highlight the improvement of our InsightEval dataset, we replicated the LLM-assisted expert annotation and scoring process on InsightBench, following the Data Quality Analysis described in the main text. Compared to the annotation of InsightEval in Table 2 of the main text, the result is presented in the Figure \ref{fig:bench_comp_llm} and \ref{fig:bench_comp_human}. These comparisons indicate that InsightEval yields enhanced quality and reliability, as corroborated by both human raters and LLM-based assessments.

Since our work is based on an optimized and refined version of the original InsightBench dataset, we have conducted a redundancy analysis on the original InsightBench dataset using the same methodology. The results are presented in Table~\ref{redundant_insightbench}. A comparison with Table~\ref{redundant} in Section~\ref{data_quality_analysis} shows that our InsightEval achieves varying degrees of improvement across nearly all metrics. This enhancement significantly strengthens the quality and credibility of our dataset.

\begin{table}[h]
  \centering
  \fontsize{9pt}{9pt}\selectfont 
  \begin{tabular}{l|ccc}
    \toprule
    Type & TC Similarity & Self‑BLEU & Distinct‑2 \\
    \midrule
    Questions & 0.0962 & 0.1746 & 0.8729 \\
    Insights & 0.1563 & 0.2120 & 0.8216 \\
    \bottomrule
  \end{tabular}
  \caption{\label{redundant_insightbench}Comparison of average TF‑IDF cosine (TC) similarity, Self‑BLEU, and Distinct‑2 diversity scores between Question and Insight entries in InsightBench.}
\end{table}

\section{Detailed Experimental Setup}
\label{exp_details}

\subsection{Models Used in this Paper}
In the construction of the dataset, the design of the agent framework, and the evaluation process, we employed several widely recognized large language models (LLMs). Detailed information regarding the versions and properties of these LLMs is provided in the Table \ref{model-details}.
\begin{table}[h!]
  \centering
  \begin{tabular}{lcc}
    \toprule
    \textbf{Models} & \textbf{Version} & \textbf{Creator} \\
    \midrule
    DeepSeek-V3      & 20250324 & deepseek \\
    GPT-4o           & 20240806 & OpenAI \\
    GPT-3.5-Turbo    & - & OpenAI \\
    O3-mini            & 20250416 & OpenAI \\
    Claude 3.7 Sonnet  & 20250219 & Anthropic \\
    Gemini 2.5 Pro     & 20250617 & Google \\
  \bottomrule
  \end{tabular}
  \caption{\label{model-details} Models used in this paper.}
\end{table}

\subsection{Agent Baselines}
\label{app_baseline_details}
We adopted a total of five agent-based baselines, with detailed descriptions provided below.  
\begin{itemize}
    \item ReAct~\cite{yao2023react}: ReAct (Reasoning and Acting) is a general paradigm that enhances LLMs' ability to solve complex tasks by tightly coupling explicit reasoning with external actions. In this framework, the model iteratively produces structured reasoning traces—supporting planning, monitoring, revision, and exception handling—and executes concrete actions that query external resources to gather necessary evidence. In our experiment, we implemented several core tools, including CSV file reading, search functionality, and numerical computation.
    \item CodeGen~\cite{majumder2025discoverybench}: The core strategy of CodeGen is to solve a task by generating the complete code in a single pass rather than performing iterative, multi-turn reasoning as in ReAct. Its workflow is as follows: the agent receives a task description, the dataset path, dataset metadata (e.g., column descriptions), and a concrete discovery goal; the prompt context supplies a code demonstration for a similar task (few-shot learning); the model then attempts to produce in one shot all code required to accomplish the task; finally, the generated code is executed and, based on the execution outputs, the model produces a natural-language hypothesis (the final answer) and a concise summary of the workflow. In the experiment, we provided the table and the target as inputs, enabling the code generation model to produce executable code for reading the table and performing analysis, which was then executed. The resulting outputs were subsequently analyzed to derive actionable insights.
    \item DeepResearchAgent~\cite{zhang2025agentorchestraorchestratinghierarchicalmultiagent}: The Deep Researcher Agent is a specialized sub-agent within the AgentOrchestra framework for large-scale, multi-round information retrieval. It can autonomously formulate optimized search queries for complex tasks and perform breadth-first retrieval across multiple search engines. Finally, it synthesizes the extracted insights into a structured research report with explicit source attributions, providing accurate and comprehensive external knowledge to support high-level planning and decision-making. Due to the absence of direct modules or tools for reading CSV tables, the table information was preprocessed and transformed into detailed textual descriptions before being provided to the agent in the experiment.
    \item Pandas Agent~\cite{langchainpandas}: A data‐science agent developed within the LangChain framework that can directly interrogate a Pandas DataFrame. In the experiment, we supply the agent with the table schema and the specified goal. Then it autonomously generates a set of questions, produces corresponding answers and insights, and finally synthesizes these insights into a summary.
    \item Agent Poirot~\cite{sahuinsightbench}: A popular recent multi‑step, multi‑round framework for insight generation published in ICLR'25. For each instance, Agent Poirot first extracts the table structure and goal, then generates an initial batch of related questions. To answer these questions, it generates Python code and executes it to obtain table information. The agent answers the corresponding questions and generates insights. Then iteratively formulates extra questions based on earlier outputs, repeating this cycle until a comprehensive set of insights is assembled.
\end{itemize}

\subsection{More Implementation Details.}
For experiments involving a single model, to enable the model to process tabular information, we first convert each table into a detailed natural-language description using a Python script in conjunction with LLM-based annotation. The resulting textual representation is then provided to the model as input. Similarly, for both single-agent and multi-agent frameworks, we equip the agents with tools that allow direct reading and querying of CSV-formatted tables. 

In addition, for the three primary backbone models used in our experiments, we set the temperature to 0 to ensure deterministic behavior. Regarding output generation, Agent Poirot is configured to run for four iterations, where each iteration produces three new questions and corresponding insights, resulting in a total of 12 generated insights. For fairness and comparability, all baseline methods are required to generate the same number of insights under identical output constraints.

\section{More Experimental Results}

\begin{figure}[h]
    \centering
    \includegraphics[width=\linewidth]{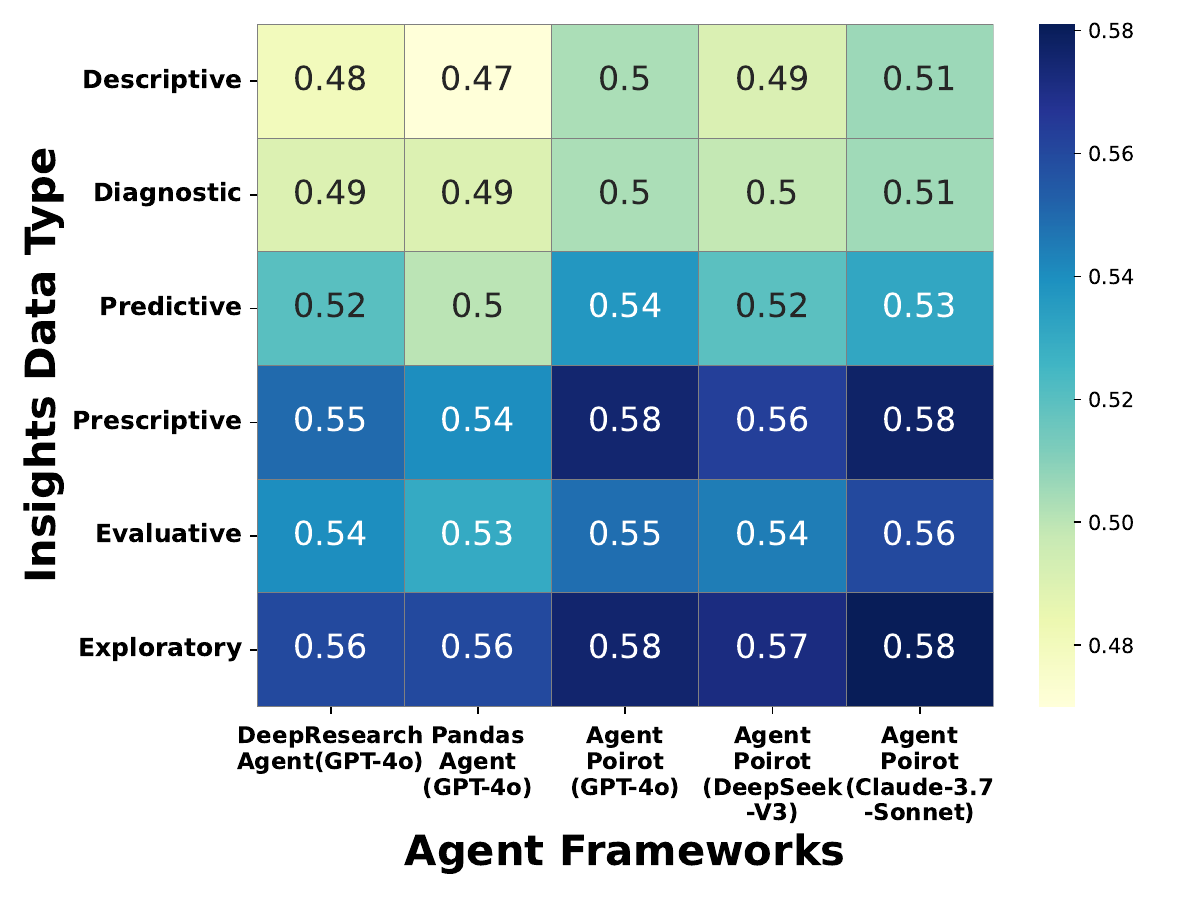}
    \caption{Agent Performance across Insight Types.}
    \label{fig:result_type}
\end{figure}

\paragraph{Agents Exhibit a Propensity for Generating Actionable and Exploratory Insights.}
As illustrated in Figure~\ref{fig:result_type}, nearly all agents achieve their highest scores within the Prescriptive, Evaluative, and Exploratory categories. Notably, under both the GPT‑4o and Claude‑3.7‑Sonnet backbones, AgentPoirot attains a peak score of 0.58 in the Prescriptive and Exploratory types. These two types respectively assess an agent's capacity for recommending executable actions and uncovering latent associations. In contrast to the other categories, the comparatively higher performance in Prescriptive and Exploratory suggests that agents are more adept at formulating practical recommendations and probing potential rules in the dataset.

\paragraph{Comprehensive Category- and Difficulty-Level Analysis Reveals Superior Performance of Agent Poirot}
We add more Experimental Results in Table \ref{claude_category_insight}, \ref{claude_novelty_category}, \ref{claude_difficulty_insight} and \ref{claude_novelty_difficulty}. Claude-3.7-Sonnet-based Agent Poirot attained the highest overall score among all agents. To further evaluate its performance, we conducted a comprehensive statistical analysis of its scores across various categories and difficulty levels. Across different performance categories, we observed that the agent achieved significantly higher Insight F1 scores in the Asset Management category. In contrast, the Novelty scores were comparatively higher in the Asset \& User Management category. Regarding difficulty levels, the agent demonstrated superior performance in Insight F1 scores at Level 1. Similarly, Level 1 yielded higher scores in terms of Novelty. Overall, the scores for the most challenging level, Level 4, were notably lower than those for other levels, suggesting that the difficulty classification is reasonably effective.

\begin{table*}[h!]
    \centering
    \setlength{\tabcolsep}{3.4pt}

    \begin{tabular}{l|cc|cc|cc}
    \toprule
    \textbf{Category}  & \multicolumn{2}{c|}{\textbf{Insights Recall}} & \multicolumn{2}{c|}{\textbf{Insights Precision}} & \multicolumn{2}{c}{\textbf{Insights $F_1$}}\\
    & ROUGE-1~~&~G-Eval & ROUGE-1~~&~G-Eval & ROUGE-1~~&~G-Eval\\
    \midrule

    Incident Management & 0.245 & 0.513 & 0.253 & 0.601 & 0.249 & 0.550 \\
    Finance Management & 0.261 & 0.572 & 0.270 & 0.654 & 0.265 & 0.608 \\
    Goal Management & 0.284 & 0.601 & 0.293 & 0.653 & 0.288 & 0.624 \\
    User Management & 0.283 & 0.589 & 0.291 & 0.645 & 0.287 & 0.614 \\
    Asset Management & \textbf{0.288} & \textbf{0.627} & \textbf{0.294} & 0.695 & \textbf{0.291} & \textbf{0.657} \\
    Strategic \& Goal Management & 0.251 & 0.446 & 0.256 & 0.553 & 0.253 & 0.492 \\
    Asset \& User Management & 0.252 & 0.485 & 0.258 & \textbf{0.727} & 0.255 & 0.582 \\
    Finance \& User Management & 0.287 & 0.616 & 0.290 & 0.673 & 0.288 & 0.643 \\
    
    \midrule
    Avg. & 0.269 & 0.556 & 0.276 & 0.650 & 0.272 & 0.596 \\

    \bottomrule
    \end{tabular}
    \caption{\label{claude_category_insight} Performance of categories on Agent Poirot based on Claude-3.7-Sonnet in Insights Level on InsightEval.}
\end{table*}
\begin{table*}[h!]
    \centering
    \setlength{\tabcolsep}{3.4pt}

    \begin{tabular}{l|cc}
    \toprule
    \textbf{Category}  & \textbf{Original Novelty Score} & \textbf{New Novelty Score}\\
    \midrule

    Incident Management & 0.647 & 0.703 \\
    Finance Management & 0.732 & 0.807 \\
    Goal Management & 0.760 & 0.802 \\
    User Management & 0.708 & 0.740 \\
    Asset Management & 0.819 & 0.833 \\
    Strategic \& Goal Management & 0.542 & 0.646 \\
    Asset \& User Management & \textbf{0.875} & \textbf{0.875} \\
    Finance \& User Management & 0.792 & 0.792 \\
    
    \midrule
    Avg. & 0.734 & 0.775 \\

    \bottomrule
    \end{tabular}
    \caption{\label{claude_novelty_category} Performance of categories on Agent Poirot based on Claude-3.7-Sonnet in Novelty on InsightEval.}
\end{table*}
\begin{table*}[h!]
    \centering
    \setlength{\tabcolsep}{3.4pt}

    \begin{tabular}{l|cc|cc|cc}
    \toprule
    \textbf{Category}  & \multicolumn{2}{c|}{\textbf{Insights Recall}} & \multicolumn{2}{c|}{\textbf{Insights Precision}} & \multicolumn{2}{c}{\textbf{Insights $F_1$}}\\
    & ROUGE-1~~&~G-Eval & ROUGE-1~~&~G-Eval & ROUGE-1~~&~G-Eval\\
    \midrule

    Level 1 & 0.253 & 0.583 & 0.255 & \textbf{0.706} & 0.254 & \textbf{0.633} \\
    Level 2 & \textbf{0.273} & \textbf{0.593} & \textbf{0.278} & 0.655 & \textbf{0.275} & 0.621 \\
    Level 3 & 0.271 & 0.583 & 0.274 & 0.667 & 0.272 & 0.619 \\
    Level 4 & 0.244 & 0.487 & 0.250 & 0.576 & 0.247 & 0.525 \\
    
    \midrule
    Avg. & 0.260 & 0.561 & 0.264 & 0.651 & 0.262 & 0.600 \\

    \bottomrule
    \end{tabular}
    \caption{\label{claude_difficulty_insight} Performance of difficulties on Agent Poirot based on Claude-3.7-Sonnet in Insights Level on InsightEval.}
\end{table*}

\section{Cases of InsightEval}
We illustrate an example of our dataset (data-8) titled Caller Incident Impact Analysis. The metadata is shown in Table \ref{DataExample}, which indicate its title, category (Incident Management), role, and difficulty (2). The table of data-8 includes 500 items about simulating ServiceNow incidents. Each item comprises structured fields such as caller id, category, state, opened at, closed at, assigned to, and priority, along with a description of the incident. The Goal of this data is to analyze incident submissions by human callers over time to detect a rising trend relative to peers.

In addition, we illustrate the details of insights and summary in Table \ref{DataExampleinsight}. Each question is annotated with a data type (Descriptive, Diagnostic, Predictive, Prescriptive, Evaluative, or Exploratory) and a synthesized insight summary.

\begin{table*}[h!]
    \centering
    \setlength{\tabcolsep}{3.4pt}

    \begin{tabular}{l|cc}
    \toprule
    \textbf{Category}  & \textbf{Original Novelty Score} & \textbf{New Novelty Score}\\
    \midrule

    Level 1 & \textbf{0.958} & \textbf{0.958} \\
    Level 2 & 0.756 & 0.795 \\
    Level 3 & 0.757 & 0.814 \\
    Level 4 & 0.591 & 0.659 \\
    
    \midrule
    Avg. & 0.766 & 0.807 \\

    \bottomrule
    \end{tabular}
    \caption{\label{claude_novelty_difficulty} Performance of difficulties on Agent Poirot based on Claude-3.7-Sonnet in Novelty on InsightEval.}
\end{table*}
\begin{table*}[ht!]
\small
\centering
\begin{tabular}{p{0.14\textwidth}|p{0.76\textwidth}}
\toprule
\textbf{Data Key}  & \textbf{Context} \\
\midrule
Header              & Caller Incident Impact Analysis (Data 8) \\
Category            & Incident Management \\
Role                & Resource Manager Analyst \\
Difficulty          & 2 \\
\midrule
Table Description   & The dataset comprises 500 entries simulating ServiceNow incidents table, detailing various attributes such as category, state, open and close dates, involved personnel, and incident specifics like description, and priority. It captures incident management activities with fields like 'opened at', 'closed at', 'assigned to', 'short description', and 'priority', reflecting the operational handling and urgency of issues across different locations and categories. \\
Table Path          & ./csvs/data-8.csv \\
User Table Path     & null \\
Table Schema        & Column: category (object)\newline  missing count: 0\newline  unique count: 5\newline  top5 unique values: ['Database', 'Hardware', 'Inquiry / Help', 'Software', 'Network']\newline Column: state (object)\newline  missing count: 0\newline  unique count: 2\newline  top5 unique values: ['Closed', 'Resolved']\newline ... \\
\midrule
Goal & Examine the distribution of incident tickets submitted by human callers by analyzing the caller id and opened at fields to evaluate frequency over time and identify any caller exhibiting a consistent upward trend in incident submissions relative to peers.\\
\bottomrule
\end{tabular}
\caption{Example of metadata in data-8 in InsightEval.}
\label{DataExample}
\end{table*}
\begin{table*}[ht!]
\small
\centering
\setlength{\tabcolsep}{2mm}
\begin{tabular}{p{0.10\textwidth}|p{0.80\textwidth}}
\toprule
\textbf{Data Key}  & \textbf{Context} \\
\midrule
Question 1 & What is the overall average number of incidents raised by callers over the recent period? \\
Data Type & Descriptive \\
Insight 1 & David Loo has raised a significantly higher number of incidents compared to other callers. \\
\midrule
Question 2 & How do the incidents raised by David Loo compare to other agents over the specific same time frame or time period? \\
Data Type & Descriptive \\
Insight 2 & David Loo's incidents are significantly higher and show a linear increasing trend over time compared to other callers. \\
\midrule
Question 3 & Are there changes in the categories of incidents raised by David Loo over time? \\
Data Type & Diagnostic \\
Insight 3 & Incidents raised by David Loo are predominantly in the Network category. \\
\midrule
Question 4 & What could be potential consequences of not addressing the anomaly or trend of raising tickets by one employee? \\ 
Data Type & Predictive \\
Insight 4 & There is a continued linear increase in ticket submissions by David Loo. \\
\midrule
Question 5 & What are the monthly counts of incident submissions for each caller and how do these counts vary over time? \\
Data Type & Descriptive \\
Insight 5 & Across the months, Bud Richman exhibited the highest submission rate peaking at 17 incidents in June 2023, while David Loo and Don Goodliffe also showed notable submission frequencies, suggesting a diverse pattern of incident reporting among different callers. \\
\midrule
Question 6 & What correlations exist between incident categories, priorities, or locations and the observed upward trend in submissions by specific callers? \\
Data Type & Diagnostic \\
Insight 6 & David Loo's submissions highlight a specific trend, as he consistently reports high-priority incidents, indicating a potential area for focused improvement within software services, particularly considering his incidents are concentrated in a critical category. \\
\midrule
Question 7 & Based on historical submission patterns, what is the forecasted number of incidents for the caller exhibiting a consistent upward trend over the next quarter? \\
Data Type & Predictive \\
Insight 7 & Notably, the trend suggests a potential increase in workload for the service team, as the caller is projected to submit an average of over 11 incidents per month, indicating rising demand for assistance. \\
\midrule
Question 8 & What operational adjustments or resource reallocations can be recommended to address and mitigate the impact of a rising trend in incident submissions by certain callers? \\
Data Type & Prescriptive \\
Insight 8 & Don Goodliffe's incident submissions peaked at 2 incidents on several occasions, suggesting a pattern of recurring issues that, if addressed early, could reduce overall ticket volume significantly. \\
\midrule
Question 9 & How complete and reliable are the opened at and caller id data fields for accurately assessing the frequency and trend of incident submissions? \\
Data Type & Evaluative \\
Insight 9 & Each caller, including identifiers such as 'ITIL User' and 'David Loo', submitted incidents uniformly, with a total of 125 incidents recorded per caller, indicating consistent engagement across the board. \\
\midrule
Question 10 & Are there any unexpected clusters or patterns in caller incident submissions that suggest emerging trends or anomalies in the data? \\
Data Type & Exploratory \\
Insight 10 & Interestingly, while all callers maintained a submission count of 125, this uniformity suggests potential saturation in incident reporting or coordination among callers, signaling an area for further investigation into their submission behaviors. \\
\midrule
Summary & The analysis shows that while several callers (like Bud Richman and Don Goodliffe) exhibit distinct submission patterns, David Loo clearly stands out. His incident submissions are significantly higher, predominantly network-related, and display a consistent, linear upward trend over time relative to his peers. This sharp increase, combined with a focus on high-priority issues, suggests that his submissions could signal emerging service challenges and a potential increase in workload for the service team. Additionally, there appears to be an overall uniformity in submission counts across some callers, indicating possible saturation or coordinated reporting behaviors that may warrant further investigation. \\
\bottomrule
\end{tabular}
\caption{Example of insights details and summary in data-8 in InsightEval.}
\label{DataExampleinsight}
\end{table*}
\begin{table*}[ht!]
\small
\centering
\setlength{\tabcolsep}{2mm}
\begin{tabular}{p{0.45\textwidth}|p{0.45\textwidth}}
    
    \toprule
    Agent Generated Output & Ground Truth \\
    \midrule
    Q1: How many employees are assigned to each manager within the IT department? & Q1: How does the number of managers and their distribution across departments affect operational effectiveness?\\
    I1: The 3:1 workload ratio between IT managers (Ed Gompf: 76 employees vs. Mariano Maury: 25 employees) suggests a potential management bottleneck that could impact employee supervision quality, project oversight, and departmental efficiency. & I1: There is a disparity in Managerial Distribution across departments.\\

    \midrule
    Q2: What is the average number of direct reports per manager across all departments compared to the IT department specifically? & Q2: What is the distribution of reportees in the IT department compared to other departments?\\
    I2: IT department managers face a management span of control nearly 4 times larger than the organizational average, which likely creates significant challenges in effective supervision, employee development, and workload management that could impact departmental performance and employee satisfaction. & I2: There is a disproportionate high number of reportees per manager in the IT Department.\\

    \midrule
    Q3: Who are the specific managers in the IT department, and what is the distribution of direct reports for each of them? & Q3: Who are the managers with the highest number of reportees?\\
    I3: The substantial workload imbalance in the IT department, where one manager (Ed Gompf) handles three times more direct reports than the other manager (Mariano Maury), could potentially lead to management inefficiencies, employee oversight issues, and burnout for the overloaded manager. & I3: There is a significant disparity among managers in terms of reportee numbers.\\

    \midrule
    Q4: Are there any location-based patterns in the assignment of IT employees to managers that might contribute to the workload imbalance? & Q4: What strategic changes can be made to improve managerial distribution and enhance departmental staffing?\\
    I4: The IT department shows a 2:1 ratio in geographic management complexity between its two managers, with Ed Gompf overseeing nearly twice as many locations (23) as Mariano Maury (12), while simultaneously managing three times as many employees (76 vs 25). & I4: The current distribution of managers and reportees within the IT department highlights significant imbalances that could impact departmental effectiveness and employee morale. To address the uneven managerial load and enhance operational effectiveness in the IT department, the organization should consider the following strategic changes: 1. Conduct a thorough review of current workload and managerial capacity to identify imbalances and potential areas of overload. 2. Implement a more equitable distribution of reportees among existing managers by reassigning team members based on current workload assessments. 3. Increase the number of managerial positions within the IT department to ensure a more manageable span of control and provide better support and oversight. 4. Offer leadership development programs to existing senior team members to prepare them for managerial roles, expanding the leadership base within the department. 5. Explore the use of technology and automation to reduce administrative burdens on managers, allowing them to focus more on leadership and strategic tasks. 6. Regularly monitor the effects of these changes on departmental performance and employee satisfaction to make further adjustments as necessary.\\

    \midrule
    Q5: What is the ratio of employees to managers in the IT department compared to other departments? & Q5: What is the distribution of reportees among managers in the IT department based on the current data?\\
    I5: The extreme disparity in management workload, with IT managers responsible for over 51 employees each compared to just 9.8 employees per manager in Customer Support, suggests a critical need for reorganization to prevent potential burnout and management inefficiencies in the IT department. & I5: Ed Gompf has a significantly larger workload compared to Mariano Maury, which may indicate a potential overload on his management responsibilities.\\

    \bottomrule

    \end{tabular}
\caption{Comparison between Agent Poirot (Claude 3.7 Sonnet) generated insights and Ground-Truth in data-27 of InsightEval (1).}
\label{tab:genvsgt}
\end{table*}

\begin{table*}[ht!]
\small
\centering
\setlength{\tabcolsep}{2mm}
\begin{tabular}{p{0.45\textwidth}|p{0.45\textwidth}}
    \toprule
    Agent Generated Output & Ground Truth \\
    
    \midrule
    Q6: What is the distribution of employee titles within the IT department, and are certain managers handling more specialized technical roles than others? & Q6: What factors or patterns in the data explain why some IT managers have significantly more reportees compared to others?\\
    I6: The IT department has a severe management imbalance with one manager overseeing 74\% of all IT staff (76 out of 103 employees), which is more than three times the workload of the other manager (25 employees), potentially creating a bottleneck in supervision and support for technical specialists. & I6: Currently, Ed Gompf manages 76 reportees, which is 304\% more than Mariano Maury's 25, highlighting a potential risk of burnout and inefficiency in the IT department due to unequal workload distribution.\\
    
    \midrule
    Q7: What is the exact number of employees assigned to each manager in the IT department, and what percentage of the total IT workforce does each manager oversee? & Q7: Based on the existing schedule and current workload distribution, how might the number of reportees per IT manager evolve in the coming month?\\
    I7: There is a significant workload imbalance in the IT department's management structure, with Ed Gompf responsible for nearly three times as many employees (76) as Mariano Maury (25), which could potentially lead to management inefficiencies and employee oversight challenges. & I7: With Ed Gompf overseeing three times more users than Mariano Maury, there is a clear risk of management overload, suggesting a need for reassessment of user allocations in the IT department.\\

    \midrule
    Q8: Are there any missing manager assignments within the IT department that could indicate organizational gaps or opportunities for better workload distribution? & Q8: What managerial adjustments or workload reallocation strategies can be implemented to mitigate potential overload in the IT department?\\
    I8: Although only 1.94\% of IT employees lack manager assignments, addressing these gaps is important as unassigned employees may lack proper oversight, potentially creating accountability issues and uneven workload distribution among existing managers in the department. & I8: Currently, Ed Gompf oversees 76 users while Mariano Maury manages only 25, indicating a substantial disparity in workload within the IT department that could lead to management challenges if not addressed.\\

    \midrule
    Q9: Which IT department managers have significantly more direct reports than the department average, and what percentage of the total IT workforce do they manage? & Q9: How reliable and complete are the 'manager' and 'department' data in identifying imbalances in workload distribution for IT managers?\\
    I9: The IT department shows a severe workload imbalance with one manager (Ed Gompf) responsible for nearly three-quarters of all IT staff, which could create a management bottleneck and potentially affect employee supervision quality and departmental efficiency. & I9: In the IT department, Ed Gompf has significantly more system users assigned to him (76) than Mariano Maury (25), highlighting a notable imbalance in workload distribution that could lead to managerial stress.\\
	
    \midrule
    Q10: What is the geographical distribution of employees reporting to each IT manager, and are there any managers responsible for teams across multiple locations? & Q10: Are there any unexpected outliers in the IT department where a manager's reported workload significantly deviates from the norm observed across other departments?\\
    I10: 100\% of IT managers oversee geographically dispersed teams, with an average of 50.5 employees per manager spread across an average of 17.5 locations, suggesting potential challenges in coordination and team management that could impact IT service delivery. & I10: This managerial workload disparity in the IT department could lead to inefficiencies and burnout, suggesting a review of user assignments to ensure a more equitable distribution among managers.\\

    \midrule
    Q11: Are there any correlations between the number of direct reports and specific employee titles or roles within the IT department that might indicate specialized workload distribution? &  \\
    I11: There is a significant workload imbalance in the IT department management structure, with Ed Gompf overseeing 76 employees while Mariano Maury manages only 25, suggesting a potential management bottleneck that could affect operational efficiency. &  \\

    \bottomrule

    \end{tabular}
\caption{Comparison between Agent Poirot (Claude 3.7 Sonnet) generated insights and Ground-Truth in data-27 of InsightEval (2).}
\label{tab:genvsgt_2}
\end{table*}

\begin{table*}[ht!]
\small
\centering
\setlength{\tabcolsep}{2mm}
\begin{tabular}{p{0.45\textwidth}|p{0.45\textwidth}}
    
    \toprule
    Agent Generated Output & Ground Truth \\
    \midrule
    Q12: What is the distribution of direct reports per manager specifically within the IT department, and which managers have significantly higher or lower numbers compared to the department average? &  \\
    I12: The 3:1 ratio of direct reports between IT managers (76 vs 25) suggests a potential management bottleneck for Ed Gompf, which could impact employee supervision quality and departmental efficiency if not addressed through workload rebalancing. &  \\

    \midrule
    Summary: The IT department shows a severe workload imbalance between its two managers, with Ed Gompf overseeing 76 employees (73.79\% of IT staff) while Mariano Maury manages only 25 employees (24.27\%), creating a 3:1 ratio that likely impacts management effectiveness. IT managers handle significantly higher workloads compared to other departments, with an average of 50.5 direct reports per manager versus the company-wide average of 13.4 reports per manager, representing a 276\% higher management burden in IT. The geographical distribution of IT staff creates additional management complexity, with Ed Gompf responsible for employees across 23 different locations while Mariano Maury oversees 12 locations, suggesting that location-based factors contribute to the workload disparity." & Summary: The insights reveal a significant imbalance in the IT department's managerial workload, with some managers overseeing far more system users than others. Specifically, Ed Gompf manages 76 users, while Mariano Maury handles only 25-a disparity that could lead to management overload, inefficiencies, and potential burnout. The recommendations include reviewing and redistributing workloads, increasing managerial positions, developing leadership among senior staff, leveraging technology to reduce administrative burdens, and continuously monitoring the impact of these changes to ensure a more equitable and effective management structure. \\

    \bottomrule

    \end{tabular}
\caption{Comparison between Agent Poirot (Claude 3.7 Sonnet) generated insights and Ground-Truth in data-27 of InsightEval (3).}
\label{tab:genvsgt_2}
\end{table*}

\section{Case Study of Comparison between Generated Results and Ground-Truth}
To enable a more intuitive comparison between the agent's output and the ground truth in our dataset, we present an illustrative example in the Table \ref{tab:genvsgt} and \ref{tab:genvsgt_2}. Notably, the agent generated 12 questions and corresponding insights, whereas the ground truth only has 10. 

According to the evaluation framework, the recall analysis shows that only the I4 ground-truth insight was unsuccessfully captured by the model outputs. For example, I2 ground-truth insight highlighted that the reported number of IT department managers was excessive, which aligns with agent-generated output I2 stating that IT managers faced an overly broad span of control. Both insights reflect a shared observation regarding managerial overload from an organizational average perspective. Besides, in the precision-based analysis, agent-generated outputs I4, I8, and I10 did not correspond to any ground-truth insights, whereas the remaining outputs exhibited strong alignment. For instance, agent-generated output I1 noted a stark 3:1 workload disparity between IT managers Ed Gompf and Mariano Maury, which closely mirrors ground-truth I6. 

Overall, our dataset effectively supports agents in uncovering meaningful insights and enables more accurate and comprehensive evaluation of their performance.

\onecolumn

\section{Prompts}
\label{appendix:prompts}
\renewcommand\lstlistingname{Prompt}

Prompt~\ref{prompt:1}, Prompt~\ref{prompt:2}, Prompt~\ref{prompt:3}, Prompt~\ref{prompt:4}, Prompt~\ref{prompt:5}, and Prompt~\ref{prompt:6} present the detailed prompts for data construction in InsightEval.

\begin{lstlisting}[breaklines=true,label={prompt:1},caption={Prompt of Goal Refinement.}]
Instruction:

Given the following table description:
<description>{table_description}</description>

Given the following table schema:
<context>{table_schema}</context>

Given the following user table scheme(if has):
<context>{user_table_schema}</context>

Given the following goal:
<goal>{goal}</goal>

Instructions:
* Analyze the given GOAL against the provided TABLE DESCRIPTION and TABLE SCHEMA. Perform these checks:
    1. Relevance: Verify if the goal can be achieved using the table's columns/relationships.
    2. Feasibility: Determine if the goal is technically achievable (e.g., required operations/joins are possible).
    3. Clarity: Assess if the goal is specific, unambiguous, and measurable.
* You need to refine the given goal by following the above checks concisely and clearly. 
* Most importantly, your new goal must be strictly enclosed within <goal></goal> tags and give your refined reason within <reason></reason>. Refer to the example response below:

Example response:
<goal>...</goal>
<reason>...</reason>

Response:
\end{lstlisting} 

\begin{lstlisting}[breaklines=true,label={prompt:2},caption={Prompt of Question Generation.}]
### Instruction:

Given the following table description:
<description>{table_description}</description>

Given the following table schema:
<context>{table_schema}</context>

Given the following user table scheme(if has):
<context>{user_table_schema}</context>

Given the following goal:
<goal>{goal}</goal>

Given the existing questions and their data type:
<questions>{exist_questions}</questions>

Instructions:
* Write a list of new questions to supplement the existing questions. New questions can be solved by the data scientists in your team to explore my table data and reach my goal.
* Your questions must be relevant to my goal. You can consider 6 types of questions in data:
    1.  **Descriptive:** Summarizes *what happened* by aggregating and visualizing historical data (e.g., plotting monthly portfolio returns).
    2.  **Diagnostic:** Explains *why it happened* by identifying correlations, patterns, and root causes (e.g., segmenting losses by asset class to find drivers).
    3.  **Predictive:** Forecasts *what is likely to happen* using statistical models on past trends (e.g., forecasting next quarter's default risk based on credit scores).
    4.  **Prescriptive:** Recommends *actions to take* to optimize outcomes or mitigate risks (e.g., suggesting portfolio rebalancing to reduce predicted volatility).
    5.  **Evaluative:** Assesses the *quality, reliability, and robustness* of the data and existing analyses/models (e.g., checking data completeness for key fields or backtesting a risk model's accuracy).
    6.  **Exploratory:** Discovers *hidden patterns, relationships, or anomalies* without predefined hypotheses (e.g., using clustering to find unexpected customer segments or identifying outlier trades).
* Considering the existing questions and their data types. Make sure that there is at least one question for each type.
* You must ask the right questions to surface anything interesting (trends, anomalies, etc.)
* Make sure these questions can realistically be answered based on the table description, table schema, and user table schema(if has).
* The insights that your team will extract will be used to generate a report.
* Each question should only have one part, that is, a single '?' at the end, which only requires a single answer.
* Do not number the questions.
* You should produce {max_questions} questions to supply. Stop generating after that.
* Most importantly, each question type and question must be enclosed within <data_type></data_type> and <question></question> tags. Refer to the example response below:

Example response:
<data_type>Descriptive</data_type>
<question>What is the distribution of the customers based on their age?</question>

### Response:
\end{lstlisting}

\begin{lstlisting}[breaklines=true,label={prompt:3},caption={Prompt of Code Generation.}]
### Instruction:

Given the goal:\n
{goal}

Given the schema:\n
{schema}

Given the data path:\n
{database_path}

Give me the Python code required to answer this question "{question}" and put a comment on top of each variable.\n\n

Instructions:
* Make a single code block for starting with ```python
* Do not produce code blocks for languages other than Python.
* Import pandas as pd, and numpy as np at the beginning.
* Save a stats JSON file that stores the data.
* Each JSON file must have a "name", "description", and "value" field that describes the data.
* If the content of the JSON file is getting too long, truncate the unnecessary parts.
* End your code with ```.

Output code:\n
\end{lstlisting}

\begin{lstlisting}[breaklines=true,label={prompt:4},caption={Prompt of Question Answering.}]
### Instruction:
You are trying to answer a question based on the information provided.

Given the following dataset schema:
<schema>{schema}</schema>

Given the goal:
<goal>{goal}</goal>

Given the question:
<question>{question}</question>

Given the data information:
<data>{data}</data>

Instructions:
* Based on the data and other information provided above, write an answer to the question enclosed with <question></question> tags.
* The answer should be a single sentence, but it should not be too high-level and should include the key details from the justification.
* Write your answer in HTML-like tags, enclosing the answer between <answer></answer> tags, followed by a justification between <justification></justification> tags.
* Refer to the following example response for the format of the answer and justification.

Example response:
<answer>This is a sample answer</answer>
<justification>This is a sample justification</justification>

### Response:
\end{lstlisting}

\begin{lstlisting}[breaklines=true,label={prompt:5},caption={Prompt of Insight Generation.}]
### Instruction:
You are trying to produce an insight based on information provided .

Given the following dataset schema:
<schema>{schema}</schema>

Given the goal:
<goal>{goal}</goal>

Given the question:
<question>{question}</question>

Given the answer:
< answer>{answer}</answer >

Instructions:
* Produce an insight that we should look into to explore my data and reach my goal.
* Write your insight in HTML-like tags, enclosing the insight between <insight></insight> tags.
* The insight should be something interesting and grounded based on the question, answer, goal, and the dataset schema, something that would be interesting. 
* The insight should be as quantiative as possible and informative and non-trivial and concise.
* The insight should be a meaningful conclusion that can be acquired from the analysis in laymans terms

Example response:
<insight>This is a sample insight</insight>

### Response:
\end{lstlisting}

\begin{lstlisting}[breaklines=true,label={prompt:6},caption={Prompt of Summary Synthesis.}]
### Instruction:
You are trying to answer a question based on information provided by a data scientist.

Given the goal:
<goal>{goal}</goal>

Given the insights list:
<insights>{insights}</insights>

Instructions:
* Given a goal, and all the insights from the above list. Give your concise summary of the above insights.
* The summary should be more about the highlights of the findings.
* Your output summary should within this tag <summary></summary>.

### Response:
\end{lstlisting}

\end{document}